\documentclass{article}

\usepackage{microtype}
\usepackage{graphicx}
\usepackage{subfigure}
\usepackage{booktabs} %

\usepackage{hyperref}

\usepackage[accepted]{icml2023}

\usepackage{amsmath}
\usepackage{amssymb}
\usepackage{mathtools}
\usepackage{amsthm}

\usepackage{graphicx}
\usepackage{amsmath}
\usepackage{amssymb}
\usepackage{booktabs}
\usepackage{pifont}
\usepackage{arydshln}
\usepackage{ulem}

\usepackage[capitalize,noabbrev]{cleveref}

\theoremstyle{plain}

\theoremstyle{definition}

\theoremstyle{remark}

\newcommand{\ie}{\textit{i}.\textit{e}., }
\newcommand{\eg}{\textit{e}.\textit{g}.}

\usepackage[textsize=tiny]{todonotes}

\begin{document}

\twocolumn[
\icmltitle{DeSRA: Detect and Delete the Artifacts of GAN-based Real-World Super-Resolution Models}

\icmlsetsymbol{equal}{*}

\begin{icmlauthorlist}
\icmlauthor{Liangbin Xie}{equal,xxx,yyy,comp1}
\icmlauthor{Xintao Wang}{equal,comp1}
\icmlauthor{Xiangyu Chen}{equal,xxx,yyy,zzz}
\icmlauthor{Gen Li}{comp2}
\icmlauthor{Ying Shan}{comp1}
\icmlauthor{Jiantao Zhou}{xxx}
\icmlauthor{Chao Dong}{yyy,zzz}
\end{icmlauthorlist}

\icmlaffiliation{xxx}{State Key Laboratory of Internet of Things for Smart City, University of Macau}
\icmlaffiliation{yyy}{Shenzhen Key Lab of Computer Vision and Pattern Recognition, Shenzhen Institute of Advanced Technology, Chinese Academy of Sciences}
\icmlaffiliation{comp1}{ARC Lab, Tencent PCG}
\icmlaffiliation{comp2}{Platform Technologies, Tencent Online Video}
\icmlaffiliation{zzz}{ Shanghai Artificial Intelligence Laboratory}

\icmlcorrespondingauthor{Chao Dong}{chao.dong@siat.ac.cn}

\icmlkeywords{Machine Learning, ICML, Real-World Image Restoration}

\vskip 0.3in
]

\printAffiliationsAndNotice{\icmlEqualContribution} %

\begin{abstract}
Image super-resolution (SR) with generative adversarial networks (GAN) has achieved great success in restoring realistic details. 
However, it is notorious that GAN-based SR models will inevitably produce unpleasant and undesirable artifacts, especially in practical scenarios. 
Previous works typically suppress artifacts with an extra loss penalty in the training phase. 
They only work for in-distribution artifact types generated during training.
When applied in real-world scenarios, we observe that those improved methods still generate obviously annoying artifacts during inference.
In this paper, we analyze the cause and characteristics of the GAN artifacts produced in unseen test data without ground-truths.  
We then develop a novel method, namely, DeSRA, to \textbf{De}tect and then ``\textbf{De}lete'' those \textbf{SR} \textbf{A}rtifacts in practice. 
Specifically, we propose to measure a relative local variance distance from MSE-SR results and GAN-SR results, and locate the problematic areas based on the above distance and semantic-aware thresholds. 
After detecting the artifact regions, we develop a finetune procedure to improve GAN-based SR models with a few samples, so that they can deal with similar types of artifacts in more unseen real data. 
Equipped with our DeSRA, we can successfully eliminate artifacts from inference and improve the ability of SR models to be applied in real-world scenarios. The code will be available at \url{https://github.com/TencentARC/DeSRA}.
\end{abstract}

\section{Introduction}
\label{sec:intro}

\begin{figure}[t]
	\begin{center}
		\includegraphics[width=\linewidth]{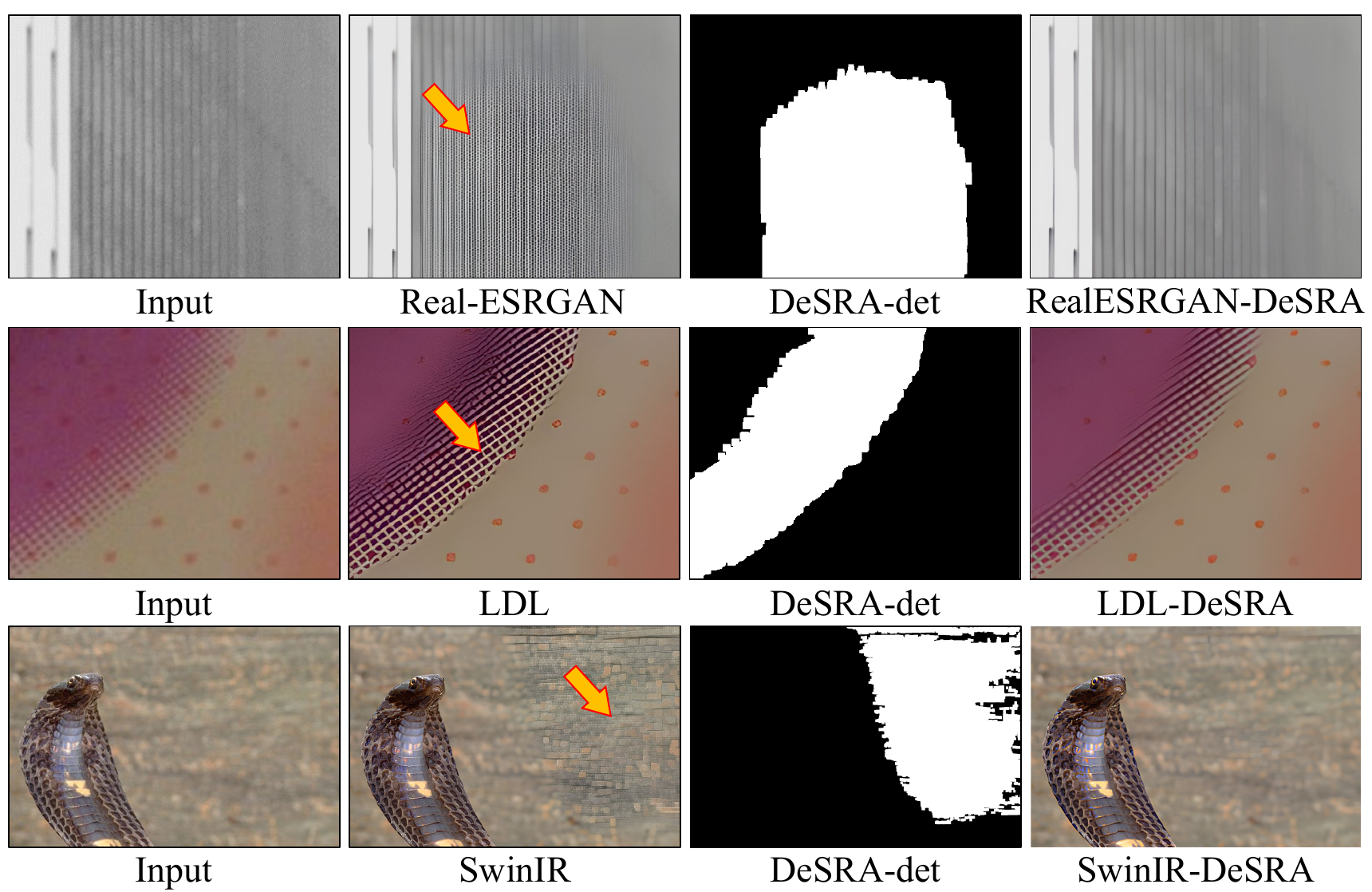}
		\vspace{-0.3cm}
		\caption{Visualization of GAN-SR artifacts, artifact detection results, and improved GAN-SR results by our proposed DeSRA. Since these GAN-inference artifacts appear on unseen real test data, existing methods that deal with GAN-SR artifacts by improving the training like LDL~\cite{ldl} would still suffer from these artifacts. Our DeSRA can effectively detect the artifact regions and then improve the SR model to eliminate those artifacts and restore visually-pleasant results. \textbf{Zoom in for best view}}
		\label{fig:teasor}
	\end{center}
	\vspace{-0.3cm}
\end{figure}

\begin{figure*}[th]
	\begin{center}
		\includegraphics[width=\linewidth]{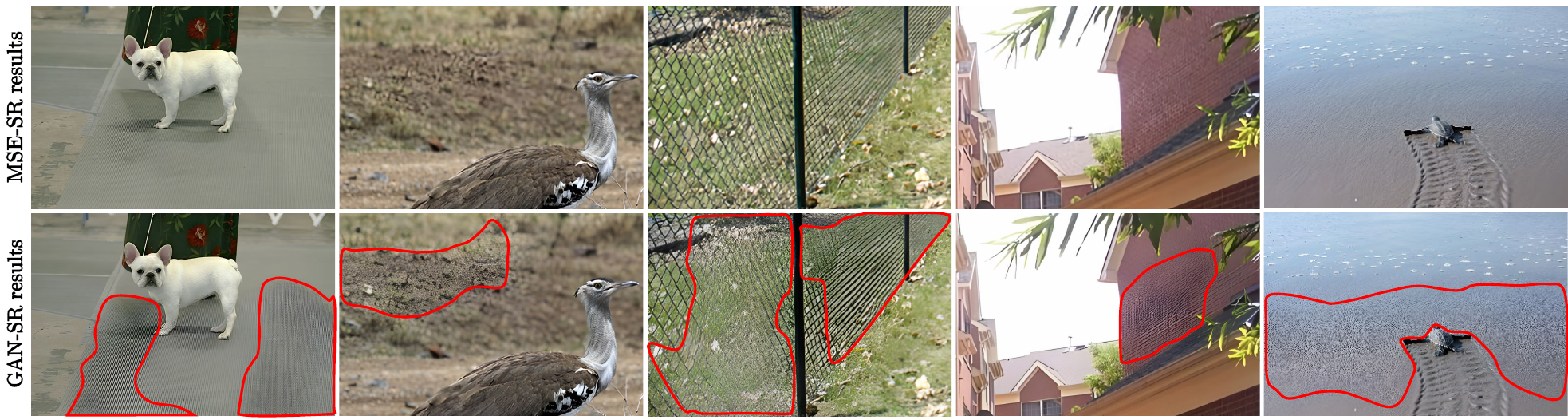}
		\vspace{-0.3cm}
		\caption{MSE-SR and GAN-SR results of some practical samples. GAN-SR results with artifacts have even worse visual quality than MSE-SR results. The artifacts are complicated with different types and characteristics, and are diverse for different image contents. Regions with red circles are GT of the detection mask.
		}
		\label{fig:artifact_examples}
	\end{center}
	\vspace{-0.3cm}
\end{figure*}

Single image super-resolution (SISR) aims to reconstruct high-resolution (HR) images from their low-resolution(LR) observations. 
Since the pioneering work of SRCNN~\cite{srcnn_eccv}, numerous approaches~\cite{edsr,rcan,swinir} have been developed and made great strides in this field. 
Among them, GAN-based methods~\cite{srgan,esrgan} have achieved great success in generating realistic SR results with detailed textures. 
Recently, BSRGAN~\cite{bsrgan} and Real-ESRGAN~\cite{realesrgan} extend GAN-based models to real-world applications and obtain promising results, demonstrating their immense potential to restore textures for real-world images. 
However, it is notorious that GAN-SR methods often generate perceptually unpleasant artifacts, which would seriously affect the user experience.
This problem is exacerbated in real-world scenarios, due to the unknown and complex degradation of LR images.

Several works~\cite{ma2020structure,ldl} have been proposed to deal with the artifacts generated by GAN-SR models. 
Typically, LDL~\cite{ldl} proposes to construct a pixel-wise map indicating the probability of each pixel being an artifact by analyzing the type of texture, and then penalizes the artifacts by adding loss during training. 
Although it indeed improves GAN-SR results, we can still observe obvious visual artifacts when inferencing real-world testing data, as shown in Fig.~\ref{fig:teasor}.  
It is hard to solve these artifacts only by improving the training on existing data pairs, since such artifacts probably do not appear during the training of GAN-SR models. 

To better illustrate this problem, we attempt to classify the GAN-SR artifacts according to the different stages they appear.
\textbf{1)} \textit{GAN-training artifacts} usually arise in the training phase, mainly due to the unstable optimization~\cite{ldl} and the ill-posed property of SR in the in-distribution data. With the presence of ground-truth images, those artifacts could be monitored during training and thus can be mitigated by improving the training, as LDL~\cite{ldl} does. 
\textbf{2)} There is another kind of artifacts that often appears in the real-world unseen data during inference, which we term as \textit{GAN-inference artifacts}. 
Those artifacts are typically out of training distribution and do not appear in the training phase.
Thus, those methods that focus on synthetic images and improve the training procedure (\eg, LDL) cannot solve those artifacts.

Dealing with GAN-inference artifacts is a new and challenging task. 
There is no ground-truth for real-world testing data with GAN-inference artifacts.
Besides, it is hard to simulate these artifacts since they may seldom or even never appear in the training set. In other words, these artifacts are unseen and out of distribution to the models.
However, solving this problem is the key to applying GAN-SR models for real-world scenarios, which has great practical value.

There are two steps to resolve the artifacts. The first step is to detect the artifact regions. In actual training of the GAN-SR model, we usually finetune it from the MSE-SR model with GAN training strategy, aiming to add fine details. Since there is no ground-truth of the inference results, we adopt the MSE-based results as the reference, which are easily accessible even for real-world data. We then design a quantitative indicator that calculates the local variance to measure the texture difference between results generated by MSE-based and GAN-based models. 
After obtaining a pixel-wise distance map, we further introduce semantic-aware adjustment to enlarge the difference in perceptually artifact-sensitive regions (\eg, building, sea) while suppressing the difference in textured regions (\eg, foliage, animal fur). 
We then filter out detection noises and perform morphological manipulations to generate the final artifact mask.

Based on the detected artifact regions, the second step is to make the pseudo GT and finetune the GAN-SR model. Firstly, we collect a small amount of GAN-based results with artifacts and replace the artifact regions with the MSE-based results according to the binarized detection masks. Then we use the combined results as the pseudo GT to construct training pairs to finetune the model for a very short period of iterations. Experimental results show that our fine-tuning strategy can significantly alleviate GAN-inference artifacts and restore visually-pleasant results on other unseen real-world data.

To summarize, \textbf{1)} We make the first attempt to analyze GAN-inference artifacts that usually appear on \textit{unseen test data without ground-truth} during inference. 
\textbf{2)} Based on our analysis, we design a method to effectively detect regions with GAN-inference artifacts. \textbf{3)} We further propose a fine-tuning strategy that only requires a small number of artifact images to eliminate the same kinds of artifacts, which bridges the gap of applying SR algorithms to practical scenarios.
\textbf{4)} Compared to previous work, our method is able to detect unseen artifacts more accurately and  alleviate the artifacts produced by the GAN-SR model in real-world test data more effectively.

\section{Related Work}
\vspace{-0.1cm}
\label{related work}

\noindent\textbf{MSE-based Super-Resolution.} 
SR methods in this category aim to restore high-fidelity results by minimizing the pixel-wise distance between SR outputs and HR ground-truth like $l1$ and $l2$ distance.
Since SRCNN~\cite{srcnn_eccv} successfully applies deep convolution neural networks (CNNs) to the image SR task, numerous deep networks~\cite{srcnn_tpami,fsrcnn,rcan,san,han,ipt,swinir,edt,hat,repsr} have been proposed to further improve the reconstruction quality.
For instance, many methods apply more elaborate convolution module designs, such as residual block~\cite{srgan,edsr} and dense block~\cite{esrgan,rdn}. 
At the same time, many works have been proposed for the Blind SR task~\cite{zhang2018learning,Gu_2019_CVPR,huang2020unfolding,Wang_2021_CVPR,faig} and video SR task~\cite{huang2017video,basicvsr,chan2022basicvsr++,liang2022vrt,shi2022rethinking,lin2022accelerating}. 
Recently, several Transformer-based networks~\cite{swinir,hat} are proposed and refresh the state-of-the-art performance.
However, due to the ill-posedness of the SR problem, optimizing the pixel-wise distance unavoidably results in smooth reconstructions that lack fine details.

\noindent\textbf{GAN-based Super-Resolution.} 
To improve the perceptual quality of SR results, GAN-based methods are proposed to introduce generative adversarial learning for SR task~\cite{srgan,esrgan,sftgan,ranksrgan,fuoli2021fourier,rad2019srobb,ma2020structure,zhang2020deep,mmsr}. 
SRGAN uses SRResNet generator and perceptual loss~\cite{johnson2016perceptual} to train the network. 
ESRGAN further improves the visual quality by adopting Residual-in-Residual Dense Block as the backbone to enhance the generator. 
To extend the GAN-SR model to real-world applications, BSRGAN~\cite{bsrgan} and Real-ESRGAN~\cite{realesrgan} design practical degradation models. 
For real-world video scenarios, RealBasicVSR~\cite{realbasicvsr} and FastRealVSR~\cite{fastrealvsr} also incorporate practical degradation models.
Despite the success, GAN-SR models often suffer from severe perceptually-unpleasant artifacts. 
SPSR~\cite{ma2020structure} proposes to alleviate the structural distortion by introducing a gradient guidance branch. 
LDL~\cite{ldl} constructs a pixel-wise map that represents the probability of each pixel being artifact and penalizes the artifacts by introducing extra loss during training.
Nonetheless, these methods would still result in artifacts in actual inference.

\section{Methodology}
\label{method}

\noindent\textbf{Preliminary: GAN-SR models} aim to learn a generative network $G$ parameterized by $\theta_{GAN}$ that estimates a high-resolution image $\hat{y}$ for a given low-resolution $x$ image as: 
\begin{equation}
    \hat{y} = G(x;\theta_{GAN}).
\end{equation}
To optimize the network parameters, a weighted combination of three sorts of losses is adopted in most GAN-SR methods~\cite{srgan,esrgan,realesrgan} as the loss function:
\begin{equation}
    \vspace{-0.1cm}
    \mathcal{L}_{GAN} = {\lambda}_{1}\mathcal{L}_{recons} + {\lambda}_{2}\mathcal{L}_{percep} + {\lambda}_{3}\mathcal{L}_{adv},
\end{equation}
where $\mathcal{L}_{recons}$ represents the pixel-wise reconstruction loss such as $l_1$ or $l_2$ distance, $\mathcal{L}_{percep}$
is the perceptual loss~\cite{johnson2016perceptual} calculating the feature distance and $\mathcal{L}_{adv}$ denotes the adversarial loss~\cite{srgan}. 
Due to the instability GAN training, in the training of a GAN-SR model, a MSE-SR model is generally trained first only using $\mathcal{L}_{reconss}$ to obtain $\theta_{MSE}$, and then the GAN-SR model is finetuned based on the pretrained $\theta_{MSE}$ using $\mathcal{L}_{GAN}$ to get the final $\theta_{GAN}$.

\subsection{Analyze GAN Artifacts Introduced in Inference}
Unlike MSE-based optimizations that naturally tend to produce over-smooth reconstruction results, GAN-based models can generate fine details benefiting from adversarial training. 
However, GAN-SR models often introduce severe perceptually-unpleasant artifacts that seriously affect the visual quality of restored images, especially in real-world scenarios. 
In some cases, the GAN-SR artifacts would make the results even worse than those generated by the MSE-based model, as shown in Fig.~\ref{fig:artifact_examples}.
Besides, these artifacts are complicated, with many types and characteristics, and are diverse for different image content.  

Essentially, methods for dealing with GAN-SR artifacts are all aimed at improving the results obtained in the inference stage.
Nevertheless, the types of artifacts that can be addressed are limited for existing methods, since they deal with the artifacts only by \textit{improving the training process}. 
For instance, LDL~\cite{ldl} processes the GAN-SR artifacts by adding penalty loss to problematic regions and improving the learning strategy.
It works for artifact types generated during the \textbf{training phase}, which exist in the in-distribution data of the training set. 
We name those artifacts as \textbf{GAN-training artifacts}. 
However, some cases of artifacts generated during the \textbf{inference phase} are out-of-distribution, namely, \textbf{GAN-inference artifacts}. 
They usually appear in unseen data without reference. 
Dealing with GAN-training artifacts would lead to better recovery of training data, but the capability of the model to process out-of-distribution data can only rely on its limited generalization ability. 
For real-world applications, how to solve more general GAN-inference artifacts is much more important. 
These artifacts are hard to synthesize during training, and thus can not be resolved by only improving the training.

In this work, we focus on processing the GAN-inference artifacts, as those artifacts have a largely negative impact on real-world applications, and solving them has great practical value. 
Due to the complexity and diversity of these kinds of artifacts, it is challenging to address all of them at once. 
We, therefore, deal with GAN-inference artifacts with the following two characteristics. 
\textbf{1)} The artifacts do not appear in the pretrained MSE-SR model (\textit{i.e.}, the generator $G$ with parameters $\theta_{MSE}$). 
\textbf{2)} The artifacts are obvious and have a large area, which can be observed at the first glance. 
Some practical examples containing such artifacts are shown in Fig.~\ref{fig:artifact_examples}.
For the former characteristic, we want to ensure that the artifacts are caused by GANs while the corresponding MSE-SR results are good references for test data to distinguish the artifacts. 
For the latter feature, we want to address those artifacts that have a great impact on visual quality.

\begin{figure}[t]
	\begin{center}
        \includegraphics[width=\linewidth]{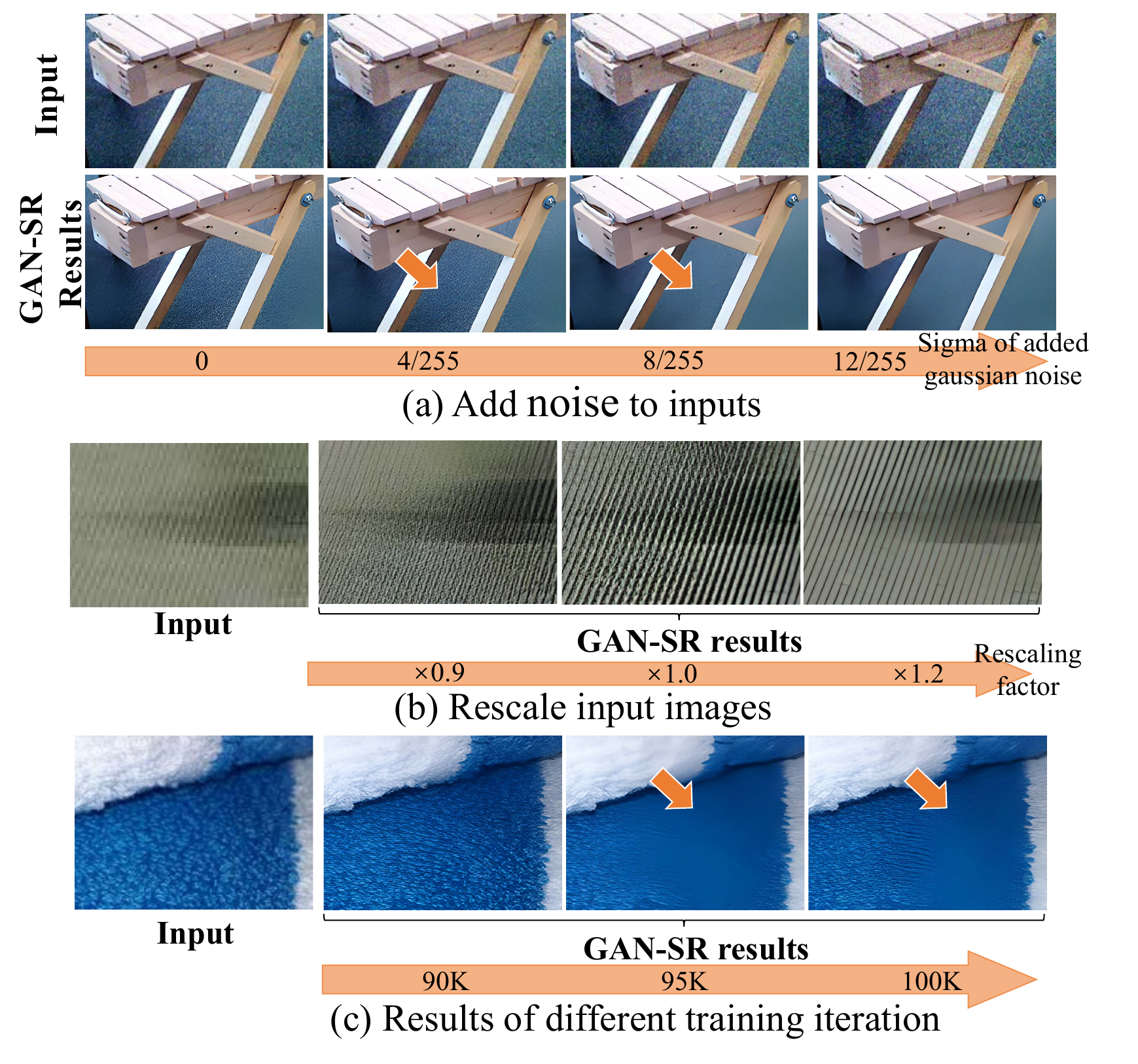}
		\caption{Observations about GAN-inference artifacts. a) Adding imperceptible Gaussian noise or b) slightly rescaling the image can alleviate the artifacts. c) Models of different iterations result in artifacts with different severity. \textbf{Zoom in for best view}}
	\label{fig:artifact_cause}
	\end{center}
\end{figure}
Before introducing the methods for addressing the artifacts, we first give a glimpse of the causes of GAN-inference artifacts. 
We found that manipulations that would slightly change the degradations, such as adding imperceptible Gaussian noise or rescaling the image, could eliminate the artifacts.
As shown in Fig.~\ref{fig:artifact_cause}, by modulating the adding noise from $\sigma=0$ to $\sigma=12/255$, the artifacts are alleviated gradually. A similar phenomenon appears when we rescale the input by setting the upscaling factor from $\times0.9$ to $\times1.2$. 
These operations essentially make the degradation of the real image close to the simulated degradations. 
This interesting observation illustrates that the reason for those GAN-inference artifacts is partly due to the out-of-distribution degradation of the input image. 
Besides, models of different training iterations also result in artifacts with different severity, as shown in Fig.~\ref{fig:artifact_cause}.
It reflects that the unstable training of GAN is also the cause of these artifacts.

\subsection{Automatically Detect GAN-inference Artifacts}
\label{detection}
At first, we want to automatically detect the regions with obvious artifacts according to some quantitative values in the inference phase before processing these artifacts. 
\textit{Due to the lack of ground-truth images, we choose the MSE-SR results as the reference to evaluate the artifacts generated by the GAN-SR model.} 
Its rationale lies in that the presentation of GAN artifacts is usually caused by too many unwanted high-frequency `details'. 
In other words, we introduce GAN training to generate fine details, but we do not want the generated content by GAN to deviate too much from MSE-SR results. 
Note that MSE-SR results are easy to access even for unseen test data, as we usually finetune the MSE-SR models to obtain GAN-SR models.
\begin{figure}[t]
	\begin{center}
		\includegraphics[width=\linewidth]{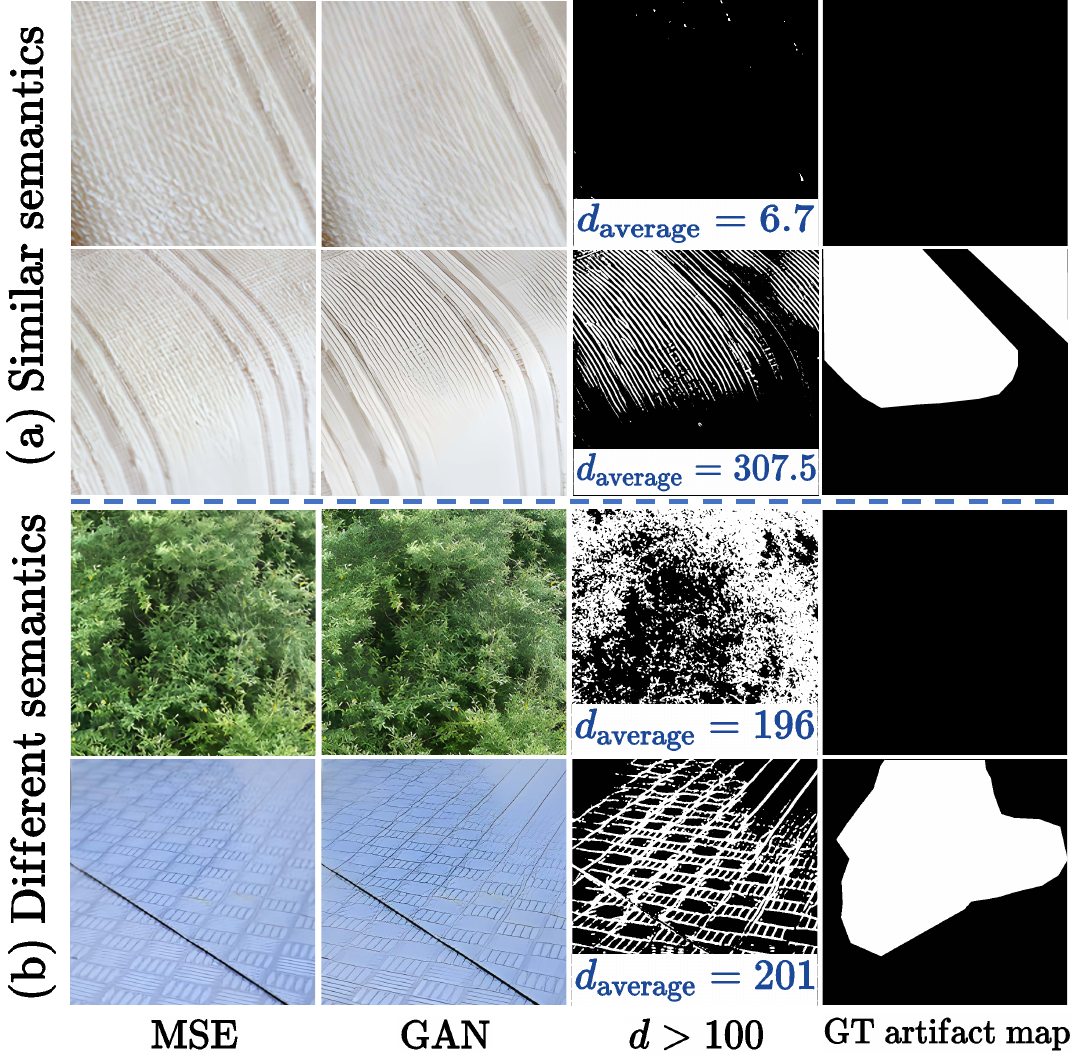}
		\caption{(a) For patches with similar semantics, too large texture difference $d$ from MSE-SR results usually indicates GAN artifacts. (b) Patches with similar $d$ have different visual quality due to their different underlying semantics. Tree regions do not have artifacts while building regions have. Thus, it is hard to measure texture differences with absolute distance. \textbf{Zoom in for best view}}
		\label{fig:relative_sigma_difference}
	\end{center}
\end{figure}

\noindent\textbf{Relative difference of local variance between MSE-SR and GAN-SR patches}.
Based on the above analysis, we propose to design a quantitative indicator to measure the difference between patches from MSE-SR and GAN-SR results as a basis for judging the artifacts. 
\begin{figure*}[!t]
	\begin{center}
		\includegraphics[width=0.9\linewidth]{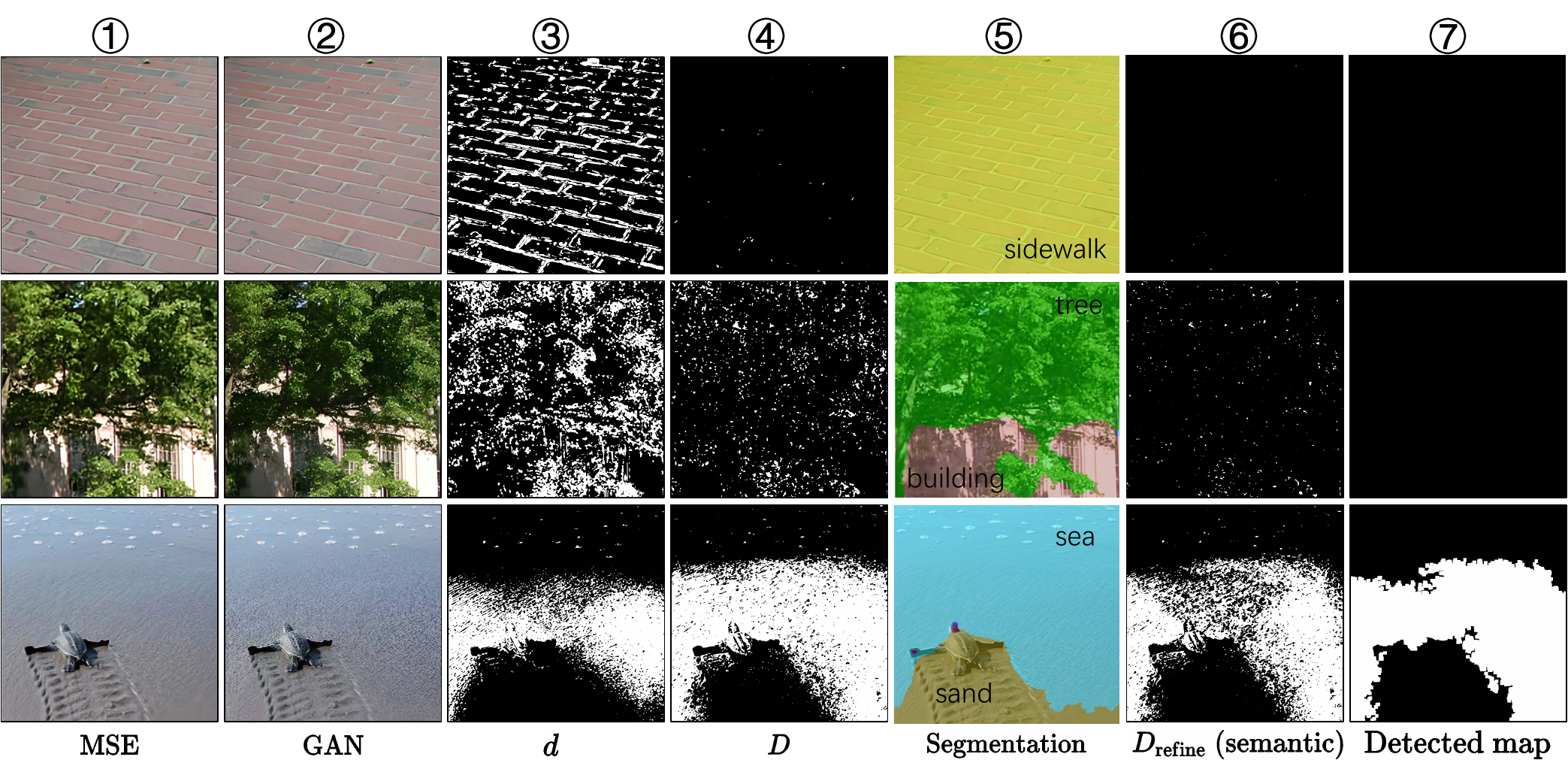}
		\caption{Visualization on the artifact detection pipeline. \ding{172} MSE-SR results. \ding{173} GAN-SR results. \ding{174} Results of directly applying texture difference $d$. \ding{175} Results of $D$ can roughly indicate artifacts but with some noises. \ding{176} Segmentation map. \ding{177} Semantic-adjusted $D$ with suppressed detection noise. \ding{178} Final detection map. \textbf{Zoom in for best view}}
		\label{fig:artifacts_detection_process}
	\end{center}
	\vspace{-0.1cm}
\end{figure*}
We adopt the standard deviation of pixel intensities within a local region $P$ to indicate the complexity of local texture as:
\begin{equation}
    \sigma(i,j)=sd(P(i-\frac{n-1}{2}:i+\frac{n-1}{2},j-\frac{n-1}{2}:j+\frac{n-1}{2})),
\end{equation}
where $\sigma(i,j)$ indicates the local standard deviation at $(i,j)$, $sd(\cdot)$ represents the standard deviation operator, and $n$ denotes the local window size and is set to 11. 
Then we calculate the difference between standard deviations of two patches to measure the texture difference $d$ as: 
\begin{equation}
    d(x,y)=(\sigma_x-\sigma_y)^2.
\end{equation}
In our case, $x$ refers to GAN-SR patches, while $y$ denotes MSE-SR patches. 
As shown in Fig.~\ref{fig:relative_sigma_difference} (a),  for patches with similar semantics, too large texture difference $d$ from MSE-SR results usually indicates GAN artifacts.
However, $d$ measures the \textbf{absolute} difference between patches, which is also related to the texture complexity itself. 
As depicted in Fig.~\ref{fig:relative_sigma_difference} (b), patches with similar $d$ have different visual quality due to their different underlying semantics. Tree regions do not have artifacts while building regions have.
Thus, we want the texture difference indicator to be a \textbf{relative} value independent of their original texture variation (\ie, the scale of $\sigma$), so we further divide $d$ by the product of $\sigma_x$ and $\sigma_y$ as:
\begin{equation}
    d^\prime(x,y)=\frac{(\sigma_x-\sigma_y)^2}{2\sigma_x\sigma_y}.
\end{equation}
To facilitate subsequent operations for the distance map, we hope to normalize the distance $d^\prime$ in the range of $[0,1]$.
Inspired by SSIM~\cite{ssim}, we adopt a similar transformation: 
\begin{equation}
    d^{\prime\prime}(x,y)=\frac{1}{1+\frac{(\sigma_x-\sigma_y)^2}{2\sigma_x\sigma_y}}=\frac{2\sigma_x\sigma_y}{\sigma_x^2+\sigma_y^2}.
\end{equation}
A constant $C$ is introduced to stabilize the division with a weak denominator. 
The final quantitative indicator can be written as:
\begin{equation}
    D=\frac{2\sigma_x\sigma_y}{\sigma_x^2+\sigma_y^2+C}. \label{D}
\end{equation}
We derive this formula step by step according to our actual needs in artifact detection, and each step has its practical meaning in our GAN-inference artifact detection. %
As shown in $3^{rd}$ and $4^{th}$ column of Fig.~\ref{fig:artifacts_detection_process}, we can observe that the generated map based on $d$ covers most of the regions with high-frequency difference between MSE-SR and GAN-SR results, but cannot distinguish these artifacts. The relative and normalized texture difference $D$ is successfully used to produce the artifacts map.

\noindent\textbf{Semantic-aware adjustment}.
After obtaining the distance map, we can exploit it to determine the regions that need to be addressed. 
However, it is not enough to only use the difference in texture complexity as a basis for judgment, because the perceptual tolerance rate of different semantic regions is different.
For example, fine details in areas with complicated textures are difficult to perceive as artifacts like foliage, hair, and \textit{etc}, while large pixel-wise differences in areas with smooth or regular textures, such as sea, sky, and buildings, are sensitive to human perception and easy to be seen as artifacts, as shown in $1^{st}$ and $2^{nd}$ column of Fig.~\ref{fig:artifacts_detection_process}.
Hence, it is required to adjust the artifact map $D$ based on the underlying semantics. 
We choose the SegFormer~\cite{segformer} as the segmentation model to distinguish different regions. Specifically, the SegFormer is trained on ADE20K, which covers most semantic concepts of the world.
To determine the reasonable adjustment weight for each class, we calculate pixel-wise $D$ values in each class of all images in the training set.
For each class, we sort all the $D$ values in descending order and set the $D$ value in the $85\%$ percentile as the adjustment weight:
\begin{equation}
    A_k=P_{85}(D_k), k\in\{1,2,...,K\},
\end{equation}
where $A_k$ is the adjustment weight for the $k^{th}$ class, $D_k$ is the $D$ value of all pixels identified as the $k^{th}$ class, and $P_{85}$ is the $85^{th}$ percentile operation. The value of $K$ is 150.
For each image, the refined detected map based on segmentation $M$ is computed as:
\begin{equation}
M(i,j)=
\begin{cases}
0, & D(i,j)/A_k \geq threshold;\\
1, & D(i,j)/A_k < threshold.
\end{cases}
\label{M}
\end{equation}
where $D(i,j)$ is the D value of pixel $(i,j)$ and $threshold$ is a hyper-parameter to control whether the current pixel is artifact or not. We empirically set the \textit{threshold} to 0.7.

We additionally perform morphological manipulations to obtain the final detected map, as shown in the $6^th$ column of Fig.~\ref{fig:artifacts_detection_process}.
Concretely, we first perform erosion using a $5\times5$ all-ones matrix. Then we implement dilation using the matrix to join disparate regions. Next, we fill the hole in the map by using a $3\times3$ all-ones matrix. Finally, we filter out discrete small regions as the detection noise.

\subsection{Improve GAN-SR Models with Fine-tuning}
\label{finetune}
\begin{figure}[ht]
	\begin{center}
		\includegraphics[width=0.9\linewidth]{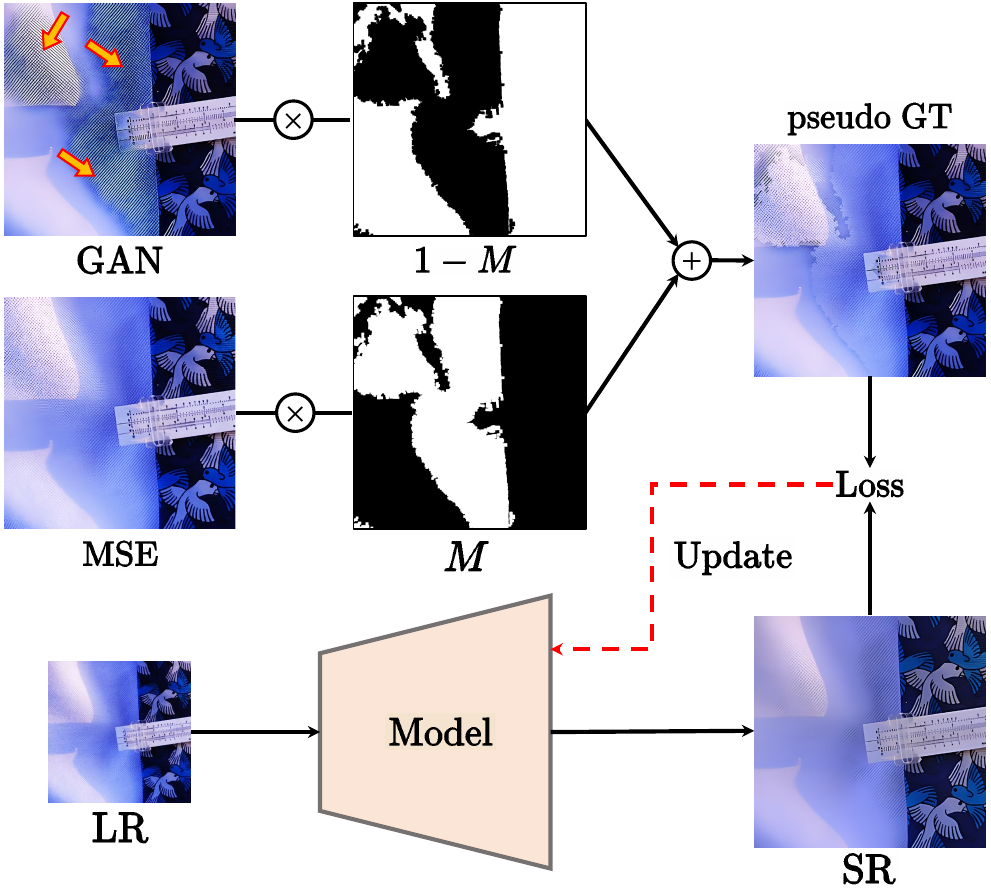}
		\caption{The procedure of our method to finetune the GAN-SR model. \textbf{Zoom in for best view}}
		\label{fig:finetune}
	\end{center}
	\vspace{-0.3cm}
\end{figure}

The detection of GAN-inference artifacts itself is of great practical value. 
We hope to further improve the GAN-SR model based on the detection results. 
Note that we aim to solve the GAN-inference artifacts for unseen real data, so there is no ground-truth for the inference results with artifacts.
In practice, ``weak restoration without artifacts is even better than strong restoration with artifacts''.  %
Thus, we exploit the MSE-SR results as the restoration reference. 

As illustrated in Fig.~\ref{fig:finetune}, we use MSE-SR results to replace the regions where artifacts were detected in GAN-SR results. The merged images serve as the pseudo GT. This process is formulated as:
\begin{equation}
    \widetilde{y}= M\cdot y_{MSE} + (1-M) \cdot y_{GAN},
\end{equation}
where $\widetilde{y}$ indicates the generated pseudo GT, $y_{MSE}$ and $y_{GAN}$ are MSE-SR and GAN-SR results, $(\cdot)$ represents the element-wise product, and $M$ is the detected artifact map. 
We then use a small amount of data to generate the data pairs $(x,\widetilde{y})$ from real data to finetune the model, where $x$ represents the LR data.
We only need to finetune the model for a few iterations (about $1K$ iterations are enough in our experiments) and the updated model would produce perceptually-pleasant results without obvious artifacts. Moreover, it does not influence other fine details in regions without artifacts.
It can effectively suppress similar kinds of artifacts in more real testing data.
The working mechanism behind this approach is that the finetuning process narrows the gap between the distribution of synthetic data and real data to alleviate the GAN-inference artifacts.

\vspace{-0.3cm}
\section{Experiments}
\label{experiments}
\vspace{-0.2cm}
\subsection{Experiment Setup}
\vspace{-0.1cm}
We exploit two state-of-the-art GAN-SR models, Real-ESRGAN~\cite{realesrgan} and LDL~\cite{ldl}, to validate the effectiveness of our method.
We use the officially released model for each method to detect the GAN-inference artifacts.
For finetuning, the training HR patch size is set to 256. The models are trained with 4 NVIDIA A100 GPUs with a total batch size of 48. We finetune the model only for 1000 iterations and the learning rate is 1e-4.

\noindent\textbf{Dataset.} Although several real-world super-resolution datasets~\cite{cai2019toward,Wang_2021_ICCV} are proposed, they assume camera-specific degradations and is far from practical scenarios. Therefore, we construct a GAN-SR artifacts dataset.
Considering the diversity of both image content and degradations, we use the validation set of ImageNet-1K~\cite{imagenet} as the real-world LR data. 
Then we choose 200 representative images with GAN-inference artifacts for each method to construct this GAN-SR artifact dataset. 
Since there is no ground-truth map for artifact regions to evaluate the algorithm, we manually label the artifact area using labelme~\cite{labelme}. 
This is the first dataset constructed for GAN-inference artifact detection. 
For the finetuning process, we further divide the dataset by using 50 pairs for training and 150 pairs for validation.

\noindent\textbf{Evaluation.}
Due to the lack of ground-truth for real-world LR data, the classic metrics such as PSNR, SSIM cannot be adopted. We also test NIQE~\cite{niqe} and MANIQA~\cite{manna}, and observe that these two metrics do not always match
perceptual visual quality~\cite{lugmayr2020ntire} (see Section~\ref{no-reference-iqa}). Thus, we consider three metrics to evaluate the detection results, including 1) \textbf{Intersection over Union (IoU)} of the detected artifact area and the ground-truth artifact area, 2) \textbf{Precision} of the detection results and 3) \textbf{Recall} of the detection results.

When using $A$ and $B$ to represent the detected artifact area and the ground-truth artifact area for a specific region $z$, IoU is given by:
\begin{equation}
    \text{IoU}=\frac{A\cap B}{A\cup B}.
\end{equation}
We can calculate IoU for each image, and we use the average IoU on the validation set to evaluate the detection algorithm. A higher IoU means better detection accuracy. 

We then define the set of regions with detected artifacts as $S$ and the set of correct samples $T$ is defined as:
\begin{equation}
    \text{T}=\{z\in S \mid \frac{A\cap B}{A} > p\}.
\end{equation}
The metric $\text{Precision}={N_T}/{N_S}$ indicates the number of correctly detected regions ($N_T$) out of the total number of detected regions ($N_S$). 
We define the set of the ground-truth regions as $G$, and the set of detected GT artifact regions $R$ is computed by:
\begin{equation}
    \text{R}=\{z\in G \mid \frac{A\cap B}{B} > p\}.
\end{equation}
The metric $\text{Recall}={N_R}/{N_G}$ represents the number of detected GT artifact regions ($N_R$) out of the total number of GT artifact regions ($N_G$). $p$ is a threshold and we empirically set it as 0.5.

\vspace{-0.2cm}
\subsection{Artifact Detection Results}
\vspace{-0.1cm}
We conduct experiments based on Real-ESRGAN~\cite{realesrgan} and LDL~\cite{ldl} to validate GAN-inference artifact detection results.
We compare our DeSRA-det described in Sec.~\ref{detection} with detection based on NIQE~\cite{niqe}, PAL4Inpainting~\cite{zhang2022perceptual}, and the modified detection protocol in LDL~\cite{niqe}.

Since there is no reference image for the unseen data in the inference phase, we choose the non-reference index NIQE~\cite{niqe} to detect the artifacts for comparing the detection scheme without using MSE-SR results. 
A similar sliding window mechanism is adopted to compute the pixel-wise map for measuring the local texture and we select the best-performing threshold for filtering the noise to obtain the final detected map. 
PAL4Inpainting~\cite{zhang2022perceptual} is a newly proposed perceptual artifacts localization method originally for inpainting. We also include it for completeness.
As the artifact detection scheme in LDL~\cite{niqe} is designed for GAN-training artifacts with ground-truth images on synthetic data, it cannot be directly applied to solve GAN-inference artifacts without GT. 
Thus, we use MSE-SR results to replace GT and set a group of threshold $\{0.001,0.005,0.01\}$ for the LDL scheme.

\begin{figure}[!t]
	\begin{center}
		\includegraphics[width=1.\linewidth]{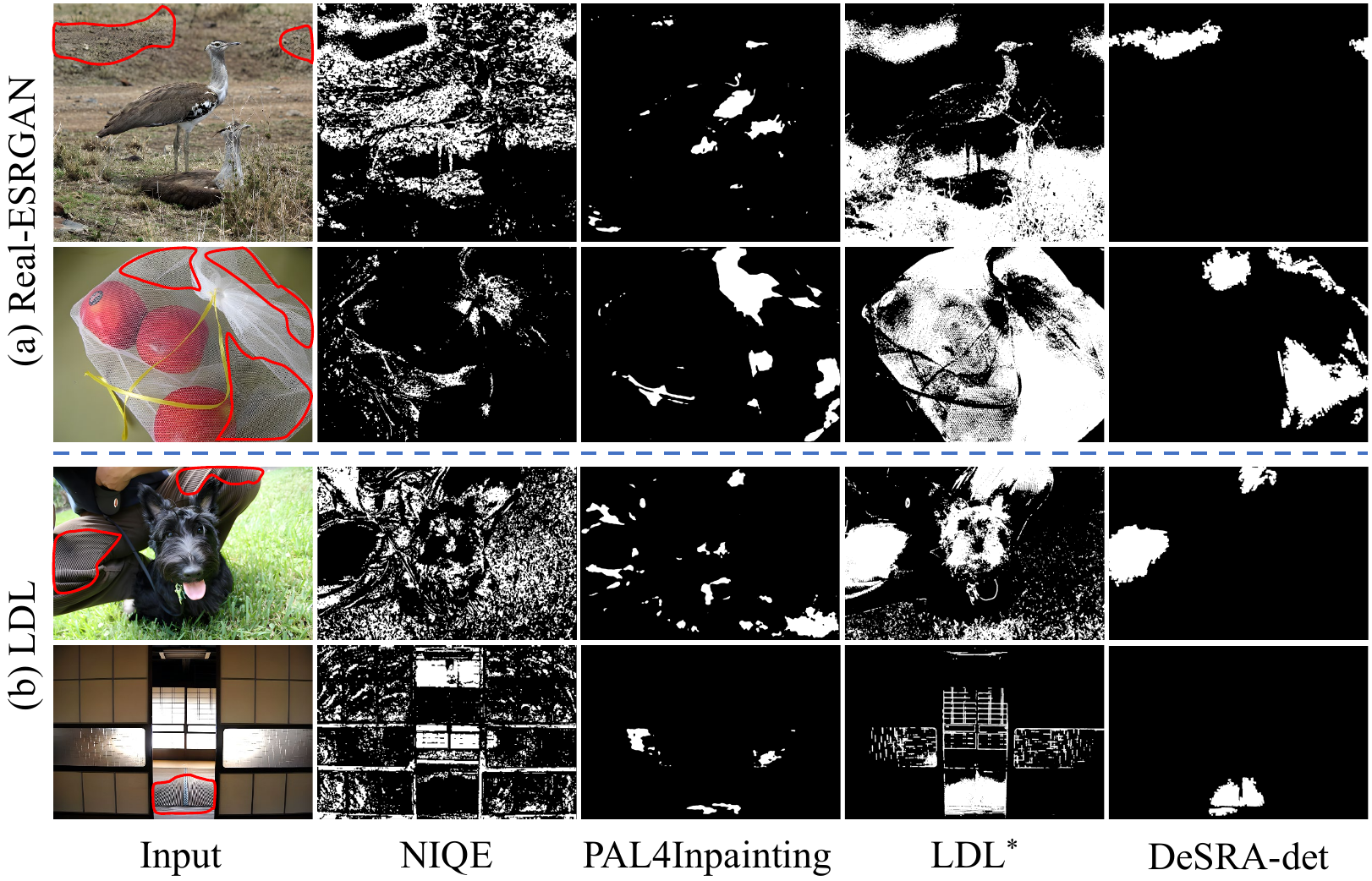}
		\vspace{-0.3cm}
		\caption{Visual comparison of different methods on artifact detection results. Regions with red circles are GT of detection mask.\textbf{Zoom in for best view}}
		\label{fig:detected_map}
	\end{center}
	\vspace{-0.3cm}
\end{figure}

Tab.~\ref{real-esrgan} shows the artifact detection results based on Real-ESRGAN. Our method obtains the best IoU and Precision that far outperform other schemes. 
Note that LDL with threshold=0.001 obtains the highest Recall. It is because this scheme treats most areas as artifacts, and thus such detection results are almost meaningless.
Similar conclusions can be drawn from Tab.~\ref{LDL} for artifact detection results based on LDL.
The visual comparison is presented in Fig.~\ref{fig:detected_map}. 
The detection results obtained by our approach have significantly higher accuracy than other schemes.

\begin{table}[t]
\center
\begin{center}
\caption{Artifact detection results based on Real-ESRGAN~\cite{realesrgan}. LDL$^\ast$ represents the modified detection method in LDL~\cite{ldl}.}
\label{real-esrgan}
\setlength{\tabcolsep}{2.1mm}{
\begin{tabular}{c|ccc} 
\hline 
Method & IoU ($\uparrow$)  & Precision & Recall\\
\hline
NIQE & 2.9 & 0.0494 & 0.1054 \\
PAL4Inpainting & 8.4 & 0.0855 & 0.0992 \\
LDL$^\ast$(threshold=0.01) & 29.9 & 0.3504 & 0.3485\\
LDL$^\ast$(threshold=0.005) & 36.2 & 0.2618 & 0.5442\\
LDL$^\ast$(threshold=0.001) & 35.3 & 0.1410 & \textbf{0.8391}\\
\textbf{DeSRA-det (ours)} & \textbf{51.1} & \textbf{0.7055} & 0.6081\\
\hline
\end{tabular}
}
\end{center}
\end{table}

\begin{table}[!t]
\center
\begin{center}
\caption{Artifact detection results based on LDL~\cite{ldl}. LDL$^\ast$ represents the modified detection method in LDL~\cite{ldl}.}
\label{LDL}
\setlength{\tabcolsep}{2.0mm}{
\begin{tabular}{c|ccc} 
\hline 
Method & IoU ($\uparrow$) & Precision & Recall\\
\hline
NIQE & 2.6 & 0.0236 & 0.1770 \\
PAL4Inpainting & 8.8 & 0.0699 & 0.1337 \\
LDL$^\ast$(threshold=0.01) & 32.7 & 0.3070 & 0.4110\\
LDL$^\ast$(threshold=0.005) & 36.7 & 0.2100 & 0.5770\\
LDL$^\ast$(threshold=0.001) & 31.1 & 0.1003 & \textbf{0.8659}\\
\textbf{DeSRA-det(ours)} & \textbf{44.5} & \textbf{0.6087} & 0.5335\\
\hline
\end{tabular}
}
\end{center}
\end{table}

\begin{figure*}[!t]
	\begin{center}
		\includegraphics[width=0.98\linewidth]{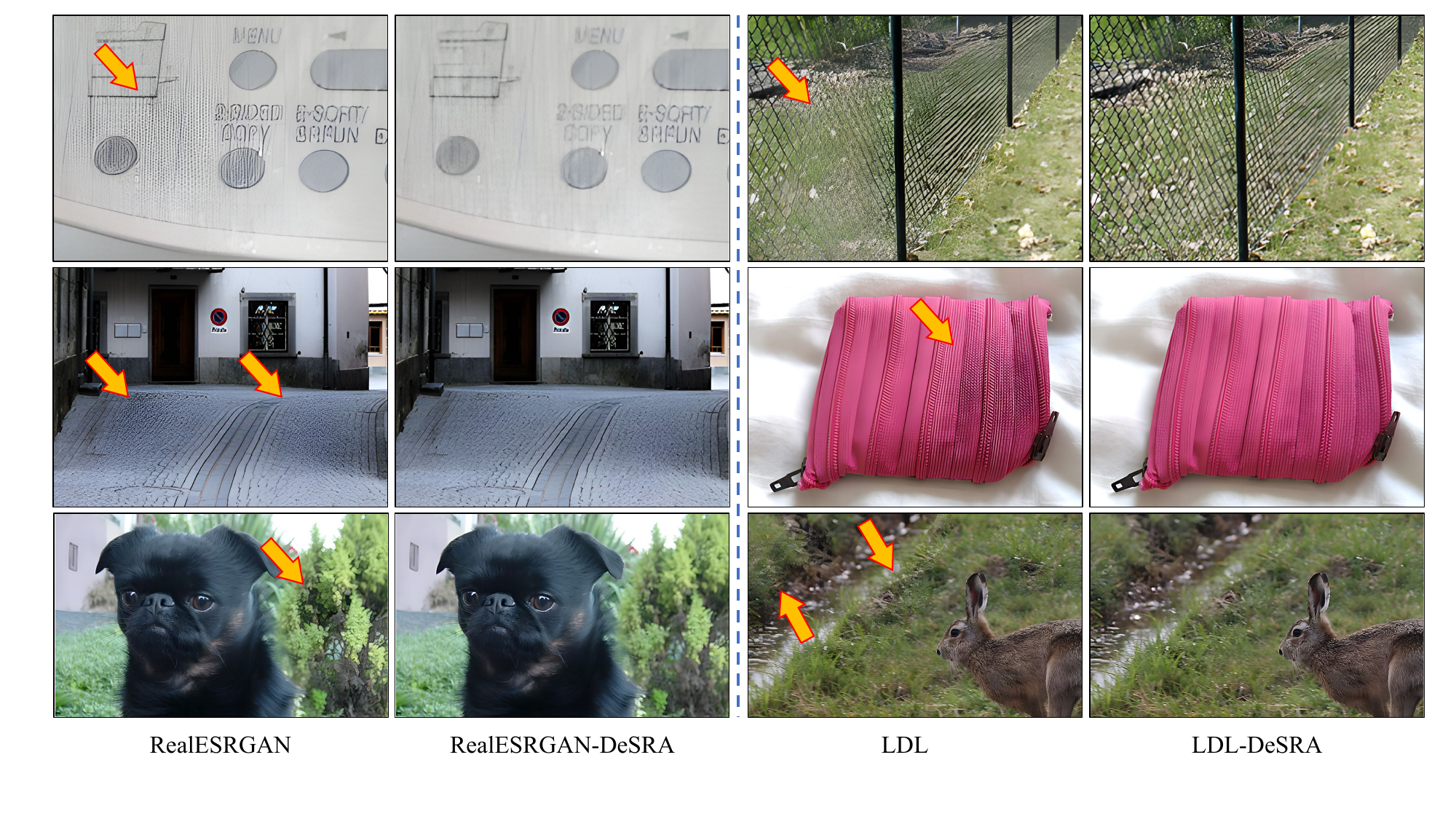}
        \vspace{-0.4cm}
		\caption{Visual comparison of results generated from original GAN-SR models and the improved GAN-SR models by using our DeSRA. Artifacts are obviously alleviated for results produced by the improved GAN-SR models. \textbf{Zoom in for best view}}
		\label{fig:qualitative_compare}
	\end{center}
	\vspace{-0.4cm}
\end{figure*}

\begin{table}[!t]
\center
\begin{center}
\caption{Artifact detection results of GAN-SR models with and without using DeSRA finetuning.}
\vspace{-0.1cm}
\label{improved}
\setlength{\tabcolsep}{0.5mm}{
\begin{tabular}{c|ccc} 
\hline 
Method & IoU ($\downarrow$) & Removal rate & Addition rate\\
\hline
Real-ESRGAN & 51.1 & - & -\\
Real-ESRGAN-DeSRA & \textbf{12.9} & 75.43\% & 0\%\\
\hdashline
LDL & 44.5 & - & -\\
LDL-DeSRA & \textbf{13.9} & 74.97\% & 0\%\\
\hline
\end{tabular}
}
\end{center}
\end{table}

\subsection{Improved GAN-SR Results}

We finetune the model to alleviate the GAN-inference artifacts based on the detected artifacts map, as described in Sec.~\ref{finetune}. 
Note that this process has a very small training cost (\textit{i.e.}, 50 training pairs with 1000 iterations). 
We compare the artifact detection results before and after using our DeSRA finetuning strategy to verify the effectiveness of improving the GAN-SR model to alleviate GAN-inference artifacts. 
The condition for judging the removal of artifacts is $A\cap B=0$, and the condition for judging the introduction of new artifacts is $A\cup B>B$.
As depicted in Tab.~\ref{improved}, after the application of our DeSRA, IoU decreases from 51.1 to 12.9 on Real-ESRGAN and from 44.5 to 13.9 on LDL, illustrating that the detected area of artifacts is greatly reduced.
The removal rate is 75.43$\%$ and 74.97$\%$, showing that three-quarters of the artifacts on unseen test data can be completely removed after finetuning.
Besides, our method does not introduce new additional artifacts, as the addition rate is 0.
We provide the visual comparison between results with and without using our method to improve GAN-SR models, as shown in Fig.~\ref{fig:qualitative_compare}.
Results generated by the improved GAN-SR models have greatly better visual quality without obvious GAN-SR artifacts compared to the original inference results.
All these experimental results demonstrate the effectiveness of our method for alleviating the artifacts and improving the GAN-SR model.

\subsection{User Study}
\begin{figure}[!t]
	\begin{center}
		\includegraphics[width=1.0\linewidth]{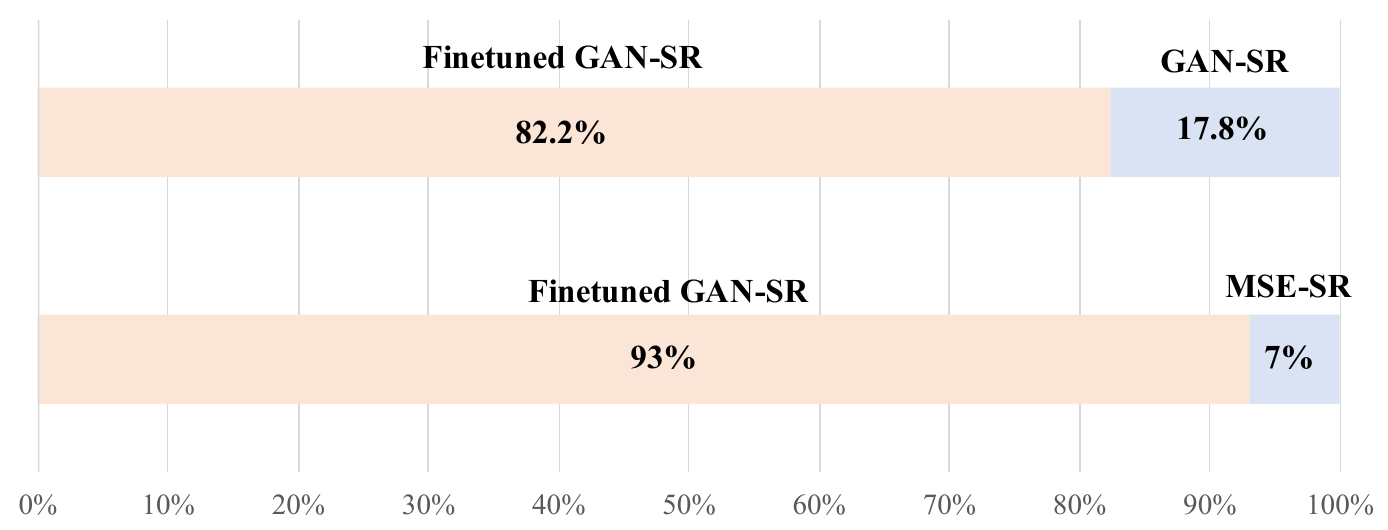}
        \vspace{-0.4cm}
		\caption{The results of user studies, comparing the results generated by the finetuned GAN-SR models with GAN-SR models and MSE-SR models. \textbf{Zoom in for best view}}
		\label{fig:user_study}
	\end{center}
	\vspace{-0.4cm}
\end{figure}

To further verify the effectiveness of our DeSRA finetuning strategy, we perform two user studies. The first is the comparison of the results generated by the original GAN-SR models and the finetuned GAN-SR models. For this experiment, the focus of comparison is on whether there are obvious artifacts. We produce a total of 20 sets of images, each containing the output results of the GAN-SR model and finetuned GAN-SR model. These images are randomly shuffled. A total of 15 people participate in the user study and select the image they think has fewer artifacts for each set. The final statistical results are shown in Fig.~\ref{fig:user_study}. 82.23\% of participants think that the results generated by fine-tuned GAN-SR models have fewer artifacts. It can be seen that our method largely removes the artifacts generated by the original model.

The second is the comparison of the results of the finetuned GAN-SR models and the original MSE-SR models. This experiment is conducted to compare whether the results generated by the model have more details. We produce a total of 20 sets of images, each containing the output results of the MSE-SR model and finetuned GAN-SR model. These images are randomly shuffled. A total of 15 people participate in the user study and select the image they think has more details for each set. The final statistical results are shown in Fig.~\ref{fig:user_study}. 93\% of participants think that the results generated by fine-tuned GAN-SR models have more details. It can be seen that the finetuned GAN-SR model generates more detailed results than the MSE-SR model.

\subsection{Ablation Study}

We first conduct the ablation study on three key designs of our artifact detection method, including  relative difference (RD) (\textit{i.e.,} from $d$ to $d^\prime$), normalization (\textit{i.e.}, from $d^\prime$ to $D$) and semantic-aware threshold. 
As shown in Tab.~\ref{ablation}, without using the relative difference suffers the lowest Precision and the full Recall. 
It is because the detection based on absolute difference would treat most areas as artifacts.
The detection scheme without normalization also results in low IoU, Precision, and Recall, since the thresholds for each sample probably have a different scale.
Using the semantic-aware threshold can improve the artifact detection results, because the sensitivity of human perception to different semantics is different. 
All these results demonstrate the necessity of the three designs in our artifact detection method.

We also conduct an ablation study for the threshold to explore its impact on artifact detection results. 
The threshold is used to control whether the pixel is the artifact or not for generating the detected map, as described in Equ.~\ref{M}.
Usually, a precision-recall curve shows the trade-off between precision and recall for different thresholds, and a high area under the curve represents both high recall and high precision.
For simplicity, we directly use ``Precision$\times$Recall'' to measure the performance of detection results to select the best threshold.
As depicted in Tab.~\ref{threshold}, the highest Precision$\times$Recall is obtained when the threshold is set to 0.7.
Thus, we select 0.7 as the default setting in our methods.

\begin{table}[!t]
\center
\begin{center}
\caption{Ablation study on the key designs for artifact detection.}
\label{ablation}
\setlength{\tabcolsep}{0.8mm}{
\begin{tabular}{ccc|ccc} 
\hline 
RD & Normalization & Semantics & IoU & Precision & Recall\\
\hline
& \checkmark & \checkmark & 16.1 & 0.0245 & 1.0000\\
\checkmark & & \checkmark & 3.6 & 0.2247 & 0.2869 \\
\checkmark & \checkmark & & 46.2 & 0.6627 & 0.5496\\
\checkmark & \checkmark & \checkmark & 51.1 & 0.7055 & 0.6081\\
\hline
\end{tabular}
}
\vspace{-0.3cm}
\end{center}
\end{table}

\begin{table}[t]
\center
\begin{center}
\caption{Influence of the threshold on the detection results.}
\label{threshold}
\setlength{\tabcolsep}{1.3mm}{
\begin{tabular}{c|cccc} 
\hline 
Threshold & IoU & Precision & Recall & Precision$\times$Recall\\
\hline
0.9 & 54.3 & 0.3255 & 0.9223 & 0.3002\\
0.8 & 56.6 & 0.5158 & 0.8123 & 0.4190\\
0.7 & 51.1 & 0.7055 & 0.6081 & \textbf{0.4290}\\
0.6 & 38.4 & 0.8343 & 0.3351 & 0.2796\\
\hline
\end{tabular}
}
\vspace{-0.3cm}
\end{center}
\end{table}

\section{Conclusion}

In this work, we analyze GAN artifacts introduced in the inference phase and propose a systematic approach to detect and delete these artifacts. We first measure the relative local variance distance from MSE-based and GAN-based results, and then locate the problematic areas based on the distance map and semantic regions. After detecting the regions with artifacts, we use the MSE-based results as the pseudo ground-truth to finetune the model. By using only a small amount of data, the finetuned model can successfully eliminate artifacts from the inference. Experimental results show the superiority of our approach for detecting and deleting the artifacts and we significantly improve the ability of the GAN-SR model in real-world applications.

\section{Acknowledgements}
This work was supported in part by Macau Science and Technology Development Fund under SKLIOTSC-2021-2023, 0072/2020/AMJ, and 0022/2022/A1; in part by Natural Science Foundation of China under 61971476, and 62276251; the Joint Lab of CAS -HK; in part by  the Youth Innovation Promotion Association of Chinese Academy of Sciences (No. 2020356).

\clearpage

\bibliography{example_paper}
\bibliographystyle{icml2023}

\newpage
\appendix
\onecolumn
\section{Appendix}

In this appendix, we provide the following materials:

\begin{enumerate}
    \item More discussions about our work. Refer to Section~\ref{discussion} in the appendix.
    \item More details of GAN-inference artifacts detection pipeline (referring to Section 3.2 in the main paper). Refer to Section~\ref{pipeline} in the appendix.
    \item More visual results of GAN-SR artifacts (referring to Section 3.1 in the main paper). Refer to Section~\ref{gan-sr-artifacts} in the appendix.
    \item Visual results of GT detection mask labeled by labelme (referring to Section 4.1 in the main paper). Refer to Section~\ref{gt-detection-mask} in the appendix.
    \item More visual comparisons of different methods on artifact detection results (referring to Section 4.2 in the main paper). Refer to Section~\ref{artifact-detection-results} in this supplementary material.
    \item Artifact detection results based on SwinIR (referring to Section 4.2 and Section 4.3 in the main paper). Refer to Section~\ref{swinir} in the appendix.
    \item More visual comparisons of results generated from original GAN-SR models and the improved GAN-SR models by using our DeSRA (referring to Section 4.2 in the main paper). Refer to Section~\ref{improved-gan-sr-desra} in the appendix.
    \item The unreliable of NIQE~\cite{niqe} and MANIQA~\cite{manna} metrics on evaluating the performance of artifacts removal (referring to Tab. 3 in the main paper). Refer to Section~\ref{no-reference-iqa} in the appendix.
\end{enumerate}

\subsection{More Discussions about Our Work}
\label{discussion}
\textbf{Discussion 1: why do we introduce the concept of GAN-inference artifacts?}

Compared with the previous work, the focus of this work is different and orthogonal. Previous works focus on improving the realness of SR results or mitigating the artifacts generated in the training phase. In real-world scenarios without ground-truth, if one algorithm can restore sharp or real textures but may also generate obviously artifacts, this algorithm is still limited in practical usage since it greatly affects the user experience. For practical application, obviously annoying artifacts are intolerable and the weak restoration results without artifacts are more acceptable by users than the strong restoration results with artifacts. Therefore, dealing with the artifacts that are generated during the inference phase, called GAN-inference artifacts, are of great value for real-world applications. Besides, some cases
of artifacts generated during the inference phase are out-of-distribution, so how to alleviate the GAN-inference artifacts is challenging and needs more attention.

\textbf{Discussion 2: why do we use MSE-SR results as the reference?}

We admit that adopting the MSE-SR results as the reference is not optimal to distinguish the GAN-inference artifacts. However, \textbf{1)} For real-world testing data, there is no ground-truth. \textbf{2)} Detecting the GAN-inference artifacts perfectly is a challenging task. From our experiments, it can be observed that when we adopt the MSE-SR results as the reference to detect the artifacts, there are many overlap areas between our detected artifact map and the GT artifact map. The quantitative and qualitative results illustrate that choosing MSE-SR results as the reference is effective for detecting the GAN-inference artifacts. Deleting the GAN-inference artifacts is a challenging task and this work is the first attempt. We believe there exist other better choices and elegant algorithms to distinguish the GAN-inference artifacts, which needs further exploration.

\textbf{Discussion 3: why do not we adopt PSNR, SSIM, NIQE $\dots$ metrics?}

\textbf{1)} The  GAN-inference artifacts appear on unseen real test data, in this circumstance, the corresponding ground-truth images are absent. Therefore, PSNR and SSIM metrics can not be adopted to evaluate the performance. \textbf{2)} We test some no-reference metrics (e.g, NIQE and MANIQA), and observe that these no-reference IQA metrics do not always match perceptual visual quality~\cite{lugmayr2020ntire} (see Section~\ref{no-reference-iqa}). 3) The focus of this work is on detecting and alleviating GAN-inference artifacts. Motivated by the binary classification task, we adopt three metrics (\eg, IOU, precision, and recall) to evaluate the performance.

\textbf{Discussion 4: why do we assume that GAN-Artifact is usually a large area?}

The GAN-inference artifacts are complicated and diverse, which appear in both large areas and small areas. Previous works focus on dealing with GAN-training artifacts and ignore the GAN-inference artifacts. When applied in real-world scenarios, those methods still generate obviously annoying artifacts during inference. Dealing with GAN-inference artifacts is a challenging task and there need several steps to resolve this problem. This work is the first attempt, and we only consider the artifacts that are obvious and have a large area, since this kind of artifact has a great impact on human perception. We hope that more researchers will pay attention to solving the GAN-inference artifacts, and the following works can deal with the GAN-inference artifacts that have a small area.

\textbf{Discussion 5: semantic segmentation.}

We admit that the detection results based on semantic segmentation are not entirely accurate, while it can get roughly accurate results to help distinguish artifacts and guide us for further processing of these artifacts, and the lost precision has a limited impact on practical applications.

\textbf{Discussion 6: online continual learning.}

Our method can provide a new paradigm combined with continual learning~\cite{de2021continual} to address the artifacts that appear in the inference stage online. 
For example, for an online SR system that processes real-world data, we can use our detection pipeline to detect whether the results have GAN-inference artifacts. We can then use the images with detected artifacts to quickly finetune the SR model, so that it can deal with similar kinds of artifacts until the system encounter a new kind of GAN-inference artifacts.  
Continual learning is widely studied on high-level vision tasks, but has not been applied to SR. 
Our approach and application scenes naturally introduce continual learning to SR.
We hope to investigate this problem in the future, since it can greatly advance the application of GAN-SR methods in practical scenarios.

\vspace{-0.2cm}
\subsection{Details of GAN-inference artifacts detection pipeline}
\label{pipeline}
In this section, we first describe the details of GAN-inference artifacts detection pipeline. Then, we provide more details about calculating the adjustment weights.

\begin{figure*}[th]
	\begin{center}
		\includegraphics[width=0.90\linewidth]{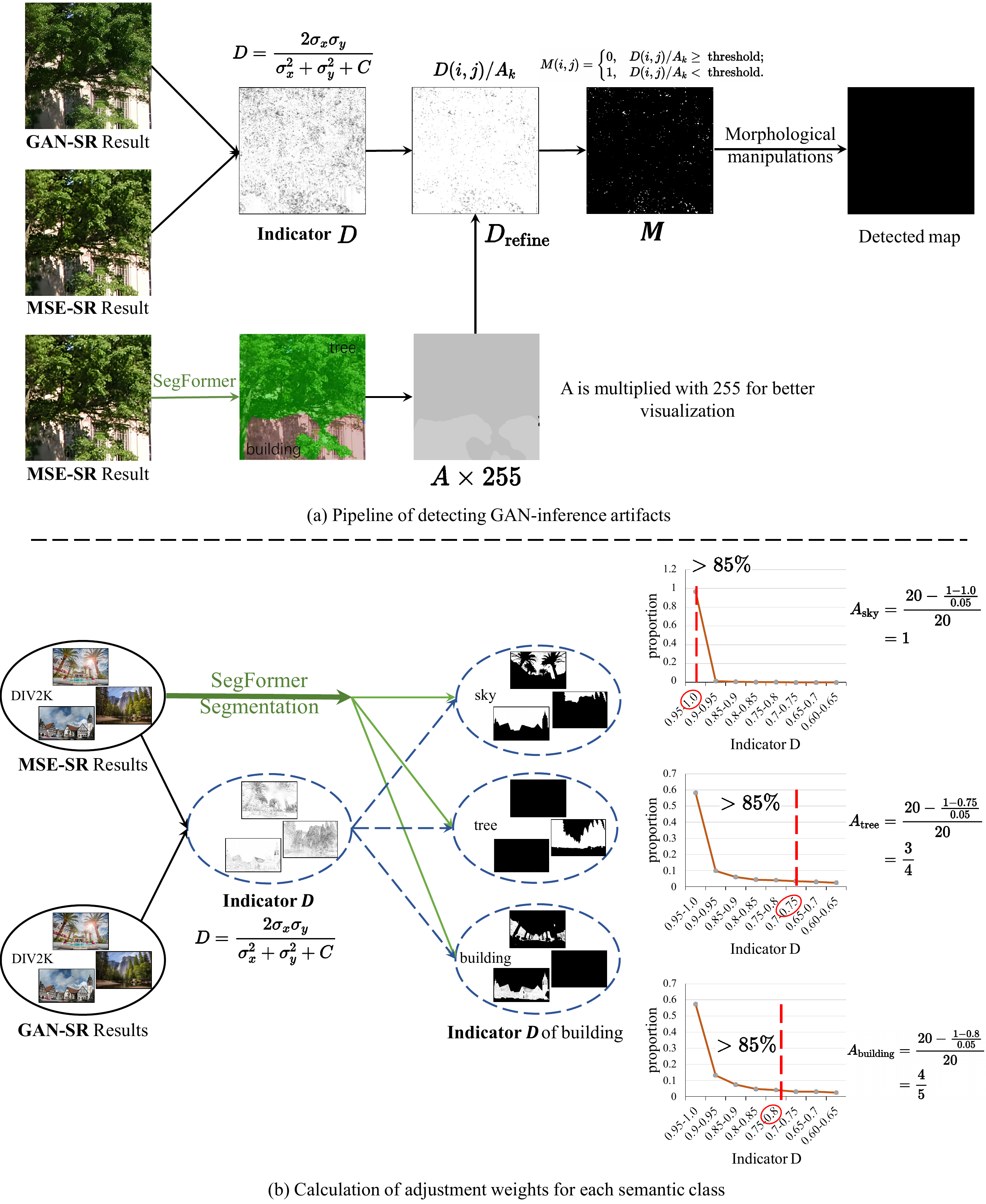}
		\caption{The overall pipeline of detecting GAN-inference artifacts and the calculation of adjustment weights.
		}
		\label{fig:detection_pipeline}
	\end{center}
\end{figure*}

\noindent\textbf{Overall pipeline of detecting GAN-inference artifacts.} The pipeline of detecting GAN-inference artifacts is shown in Fig.~\ref{fig:detection_pipeline} (a). For a GAN-SR and MSE-SR result, we first calculate the indicator $D$ according to equation 7 in the main paper. Then we generate the segmentation map of MSE-SR result by adopting SegFormer. The segmentation map will be converted into semantic-aware adjustment weight $A$ according to the calculated adjustment weights of each semantic class (Fig.~\ref{fig:detection_pipeline} (b)). By combining $A$, $D$ and setting $threshold$, we can obtain the refined detected map $M$:
\begin{equation}
M(i,j)=
\begin{cases}
0, & D(i,j)/A_k \geq threshold;\\
1, & D(i,j)/A_k < threshold,
\end{cases}
\label{M}
\end{equation}
where $D(i,j)$ is the D value of pixel $(i,j)$ and $threshold$ is empirically set to 0.7. At last, we perform morphological manipulations to obtain the final detected map. Concretely, we first perform erosion using a $5\times5$ all-ones matrix. Then we implement dilation using the matrix to join disparate regions. Next, we fill the hole in the map by using a $3\times3$ all-ones matrix. Finally, we filter out discrete small regions as the detection noise. 

Note that the visualization results of indicator $D$ and $D_{refine}$ in Fig.~\ref{fig:detection_pipeline} (a) are different from Fig. 5 in the main paper. Here we show their original values. In the main paper, for better understanding, we show their corresponding binary maps by comparing their original values with the threshold $0.7$. Values that are smaller than the threshold $0.7$ are set to $1$. 

\noindent\textbf{Details of calculating adjustment weights.} The calculation of adjustment weights for each semantic class is illustrated in Fig.~\ref{fig:detection_pipeline} (b). We first generate the corresponding low-resolution version by adopting the degradation model used in the training phase on the DIV2K training dataset. 
Then, we generate the MSE-SR and GAN-SR results for each distorted image. After that, we calculate pixel-wise indicator $D$ between the MSE-SR and GAN-SR results. To distinguish $D$ of each semantic class, we choose SegFormer~\cite{segformer} as the segmentation model, and obtain the segmentation map of MSE-SR results. 
By incorporating the segmentation map and indicator $D$, we get pixel-wise $D$ values in each class of DIV2K. For each class, we sort all the D values in descending order and set the $D$ value in the $85\%$ percentile as the adjustment weight:
\begin{equation}
    A_k=P_{85}(D_k), k\in\{1,2,...,K\},
\end{equation}
where $A_k$ is the adjustment weight for the $k^{th}$ class, $D_k$ denotes the $D$ value of all pixels identified as the $k^{th}$ class, and $P_{85}$ is the $85^{th}$ percentile operation. For example, the values of $A_{\mathrm{sky}}$, $A_{\mathrm{tree}}$ and $A_{\mathrm{building}}$ are $1$, $0.75$ and $0.80$, respectively. 

\clearpage
\clearpage
\newpage
\subsection{More Visual Results of GAN-SR Artifacts}
\label{gan-sr-artifacts}
\begin{figure*}[h]
	\begin{center}
		\includegraphics[width=\linewidth]{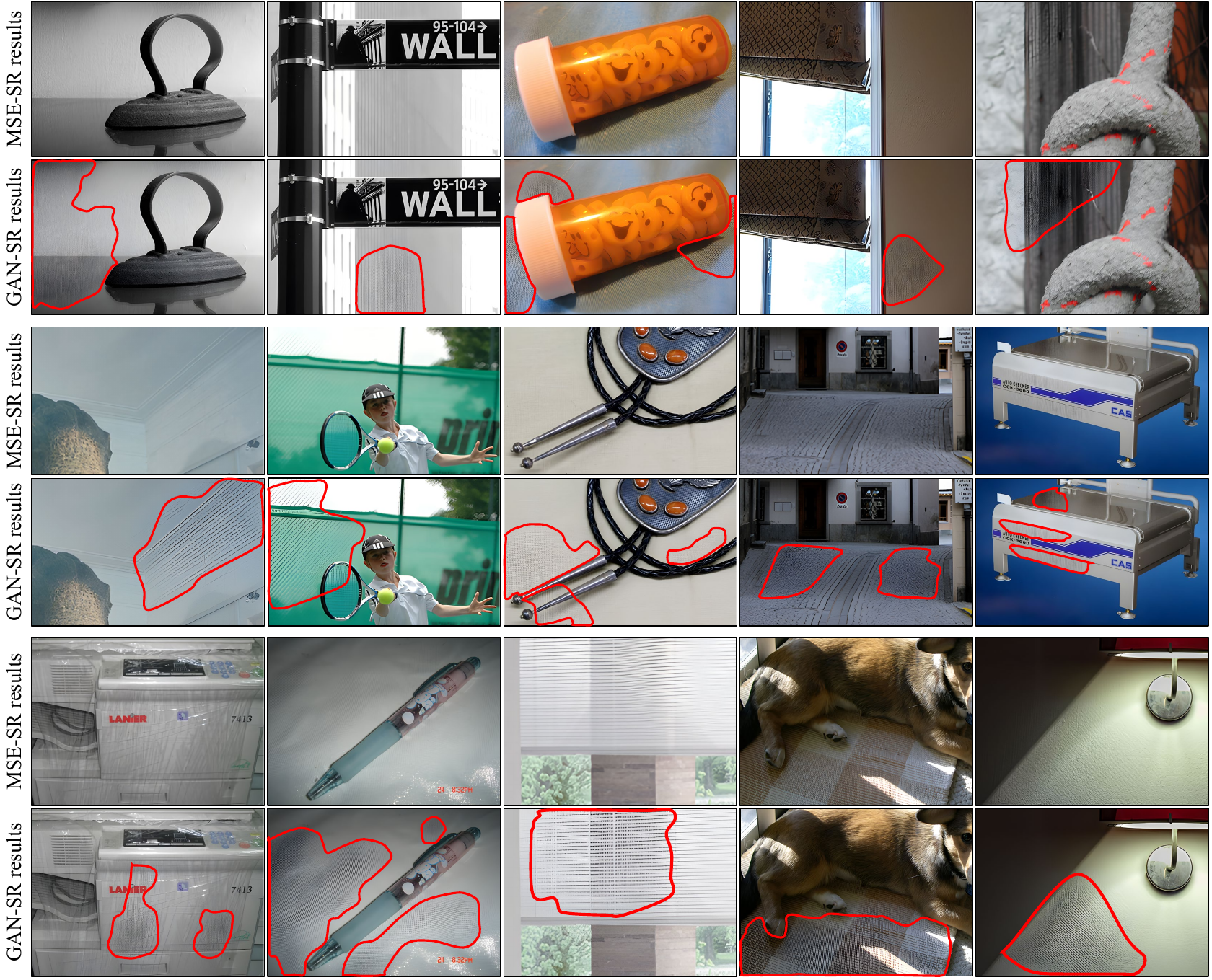}
		\caption{MSE-SR and GAN-SR results of some practical samples. GAN-SR results with artifacts have even worse visual quality than MSE-SR results. The artifacts are complicated with different types and characteristics, and are diverse for different image contents. Regions with red circles are GT of the detection mask. \textbf{Zoom in for best view}
		}
		\label{fig:artifact_examples_supp}
	\end{center}
\end{figure*}

In real-world scenarios, GAN-SR models often introduce severe perceptually-unpleasant artifacts that seriously affect the visual quality of restored images. As depicted in Fig.~\ref{fig:artifact_examples_supp}, in some cases, the GAN-SR artifacts would make the results even worse than those generated by the MSE-based model.

\clearpage
\clearpage
\newpage
\subsection{Visual Results of GT Detection Mask}
\label{gt-detection-mask}
\begin{figure*}[th]
	\begin{center}
		\includegraphics[width=\linewidth]{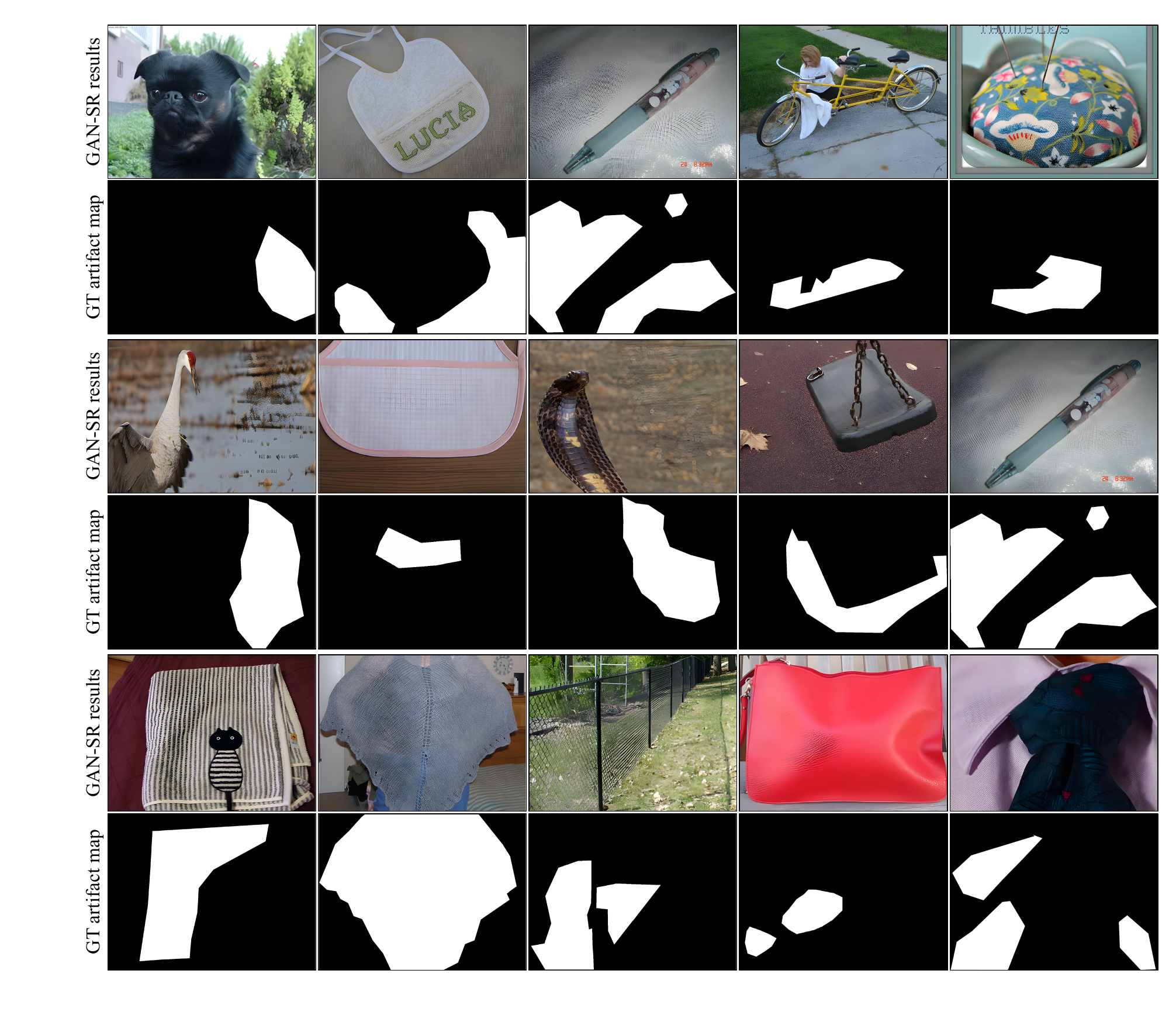}
		\caption{Visualization of GT artifact map. For GAN-SR results with artifacts, we manually generate their corresponding binary artifact map with labelme. The white regions of GT artifact map indicate the artifact regions in GAN-SR results.}
		\label{fig:gt_detection_mask}
	\end{center}
\end{figure*}

For Real-ESRGAN~\cite{realesrgan}, LDL~\cite{ldl} and SwinIR~\cite{swinir}, we construct their independent GAN-SR artifacts datasets. Each dataset contains $200$ representative images with GAN-inference artifacts. Since there is no ground-truth map for artifact regions to evaluate the algorithm, we manually label the artifact areas using labelme~\cite{labelme} and generate a binary map to indicate the artifact region, as shown in Fig.~\ref{fig:gt_detection_mask}.

\clearpage
\newpage

\subsection{More Visual Comparisons of Different Methods on Artifact Detection Results}
\label{artifact-detection-results}

\begin{figure*}[th]
	\begin{center}
		\includegraphics[width=0.92\linewidth]{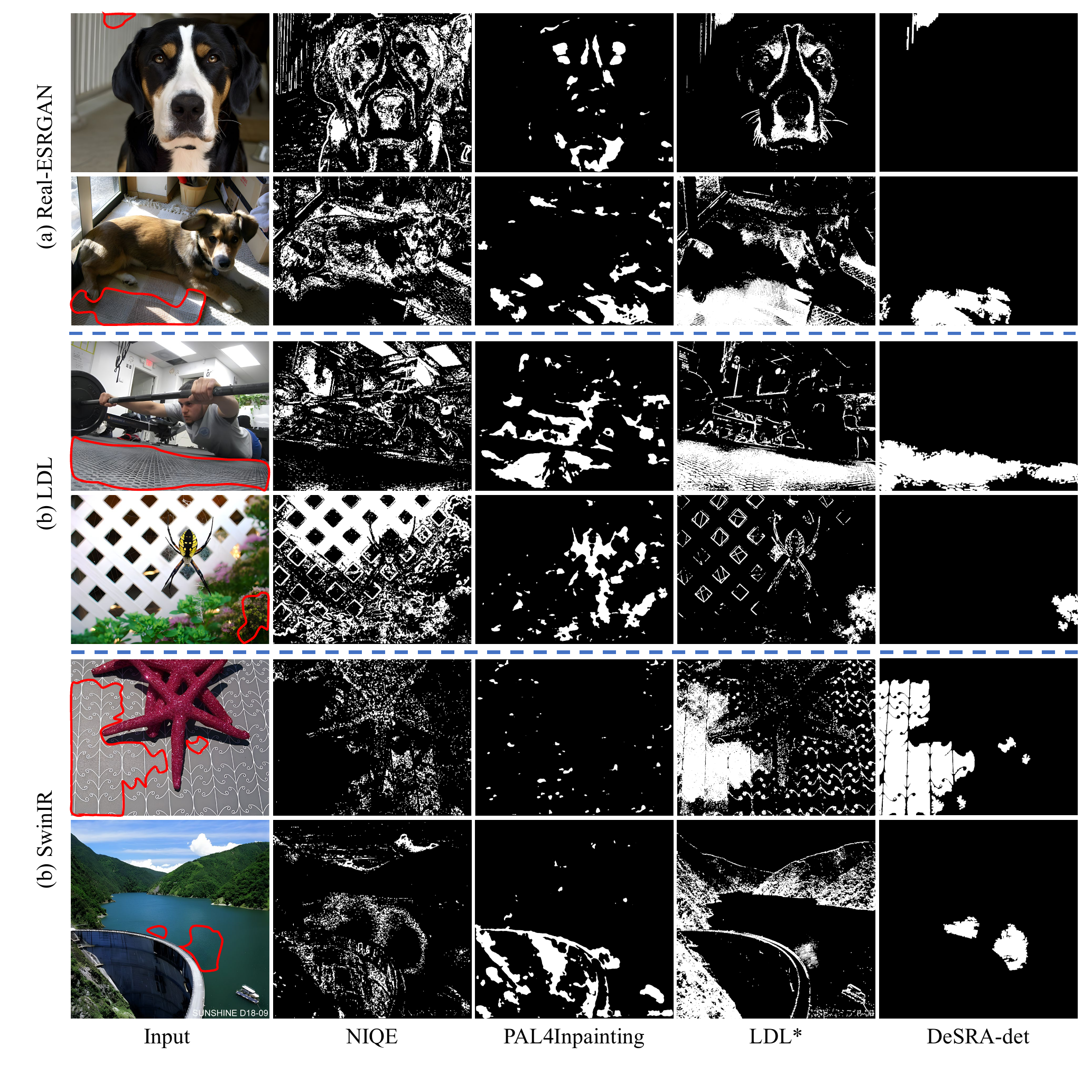}
		\caption{Visual comparison of different methods on artifact detection results. Regions with red circles are GT of the detection mask. \textbf{Zoom in for best view}}
		\label{fig:artifact_detection}
	\end{center}
\end{figure*}

For the GAN-inference artifacts generated by Real-ESRGAN~\cite{realesrgan}, LDL~\cite{ldl} and SwinIR~\cite{swinir}, we compare different methods on artifact detection results. The visual comparison is presented in Fig.~\ref{fig:artifact_detection}.
The detection results obtained by our approach have significantly higher accuracy than other schemes.

\clearpage
\newpage

\subsection{Artifact Detection Results based on SwinIR}
\label{swinir}
To validate the effectiveness of our proposed GAN-inference artifact detection algorithm and fine-tuning strategy, we further conduct experiments based on SwinIR. Due to the lack of the official-released pretrained weight of the discriminator, we retrain SwinIR using the officially released codes\footnote{https://github.com/JingyunLiang/SwinIR} in real setting and obtain the corresponding MSE-SR and GAN-SR models. For the GAN-inference artifacts generated by SwinIR, the artifact detection results are shown in Tab.~\ref{tab:swinir}. We can observe that our method obtains the best IoU and Precision that far outperform other schemes. 

After obtaining the detected artifacts map, we finetune SwinIR with $1000$ iterations to alleviate the GAN-inference artifacts. As depicted in Tab.~\ref{tab:improved-swinir}, after the application of our DeSRA, IoU decreases from 57.9 to 21.8, illustrating that the detected area of artifacts is greatly reduced. The removal rate is 61.35\%, showing that three-fifths of the artifacts on unseen test data can be completely removed after fine-tuning. Besides, our method does not introduce new additional artifacts, as the addition rate is 0.

\begin{table*}[th]
\center
\begin{center}
\caption{Artifact detection results based on SwinIR~\cite{swinir}. LDL$^\ast$ represents the modified detection method in LDL~\cite{ldl}.}
\label{tab:swinir}
\setlength{\tabcolsep}{10mm}{
\begin{tabular}{c|ccc} 
\hline 
Method & IoU ($\uparrow$)  & Precision & Recall\\
\hline
NIQE~\cite{niqe} & 2.3 & 0.0227 & 0.0668 \\
PAL4Inpainting~\cite{zhang2022perceptual} & 6.3 & 0.0547 & 0.1277 \\
LDL$^\ast$(threshold=0.01)~\cite{ldl} & 11.0 & 0.3039 & 0.1176 \\
LDL$^\ast$(threshold=0.005) & 19.4 & 0.2377 & 0.2647 \\
LDL$^\ast$(threshold=0.001) & 29.2 & 0.1473 & 0.7380\\
LDL$^\ast$(threshold=0.0001) & 29.2 & 0.1473 & 0.7380\\
\textbf{DeSRA-det (ours)} & \textbf{57.9} & \textbf{0.7600} & \textbf{0.7412}\\
\hline
\end{tabular}}
\end{center}
\end{table*}

\begin{table}[th]
\center
\begin{center}
\caption{Artifact detection results of SwinIR~\cite{swinir} with and without using DeSRA finetuning.}
\label{tab:improved-swinir}
\setlength{\tabcolsep}{8mm}{
\begin{tabular}{c|ccc} 
\hline 
Method & IoU ($\downarrow$) & Removal rate & Addition rate\\
\hline
SwinIR~\cite{swinir} & 57.9 & - & -\\
SwinIR-DeSRA & \textbf{21.8} & 61.35\% & 0\%\\
\hline
\end{tabular}
}
\end{center}
\end{table}

\subsection{More Visual Comparisons between the Original GAN-SR Models and the Improved GAN-SR Models with DeSRA}
\label{improved-gan-sr-desra}

\begin{figure*}[th]
	\begin{center}
		\includegraphics[width=0.92\linewidth]{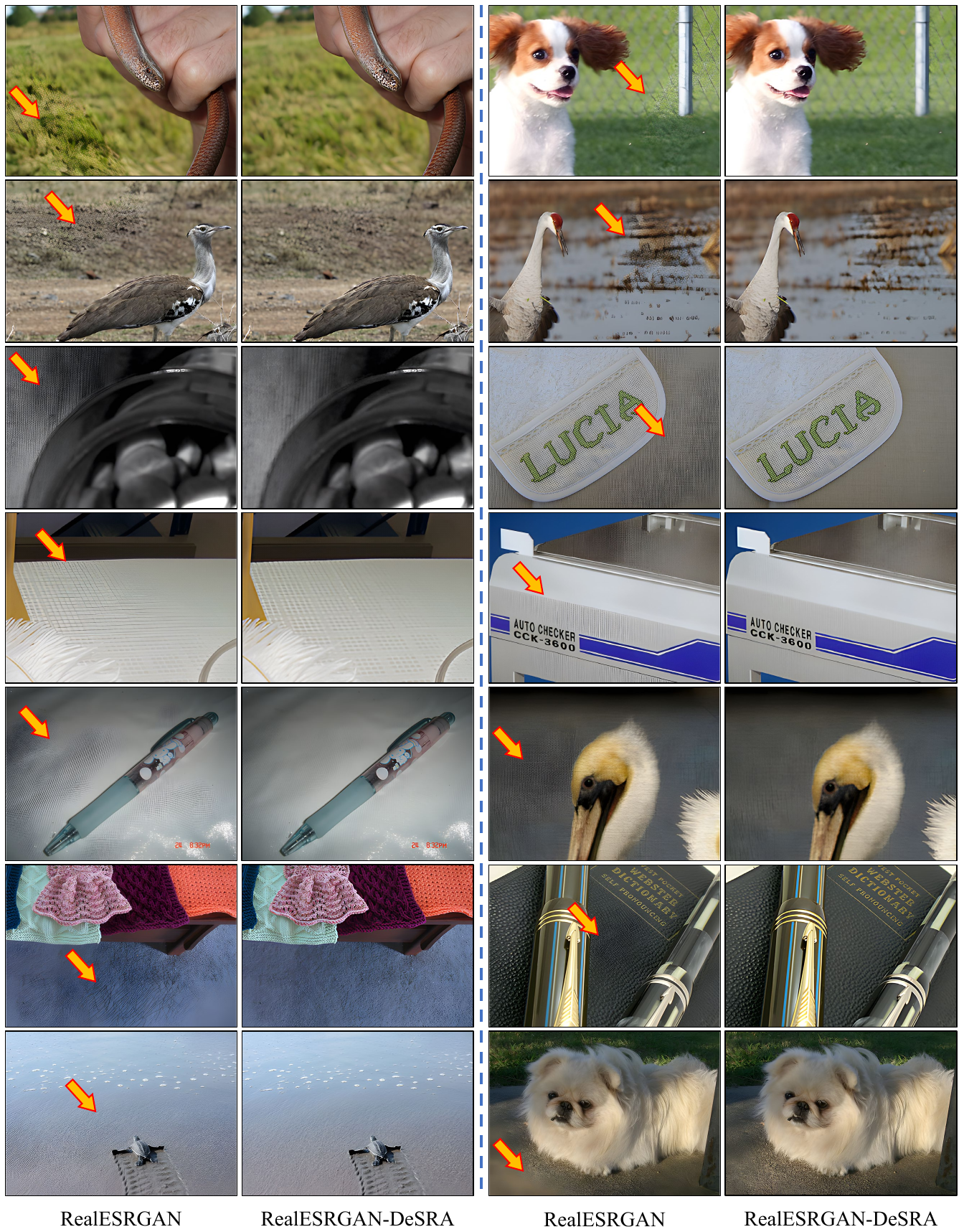}
		\caption{Visual comparison of results generated from RealESRGAN and RealESRGAN-DeSRA. Artifacts are obviously alleviated for results produced by RealESRGAN-DeSRA.}
		\label{fig:realesrgan_finetune}
	\end{center}
\end{figure*}

\begin{figure*}[th]
	\begin{center}
		\includegraphics[width=0.88\linewidth]{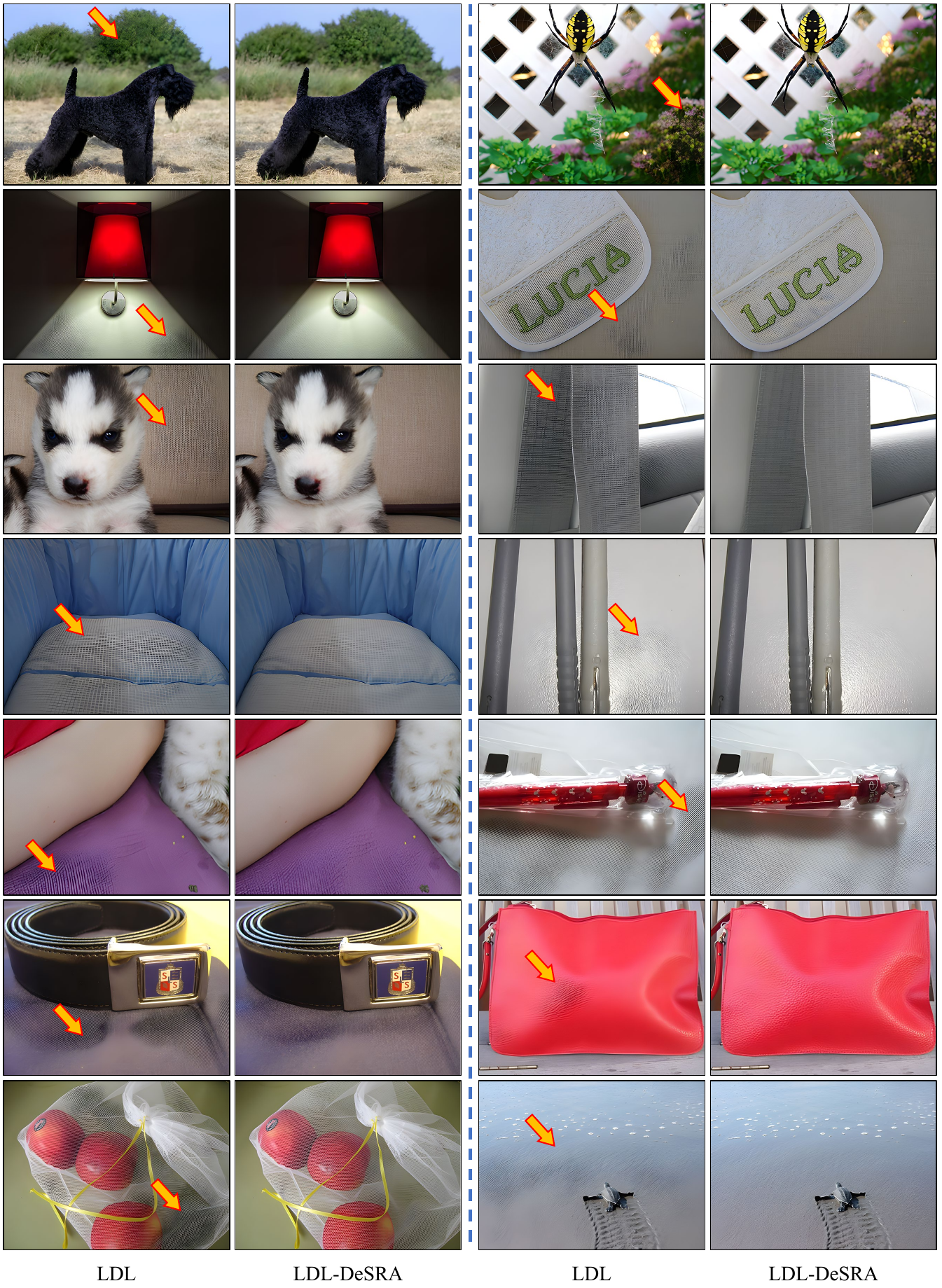}
		\caption{Visual comparison of results generated from LDL and LDL-DeSRA. Artifacts are obviously alleviated for results produced by LDL-DeSRA. \textbf{Zoom in for best view}}
		\label{fig:ldl_finetune}
	\end{center}
\end{figure*}

\begin{figure*}[th]
	\begin{center}
		\includegraphics[width=0.88\linewidth]{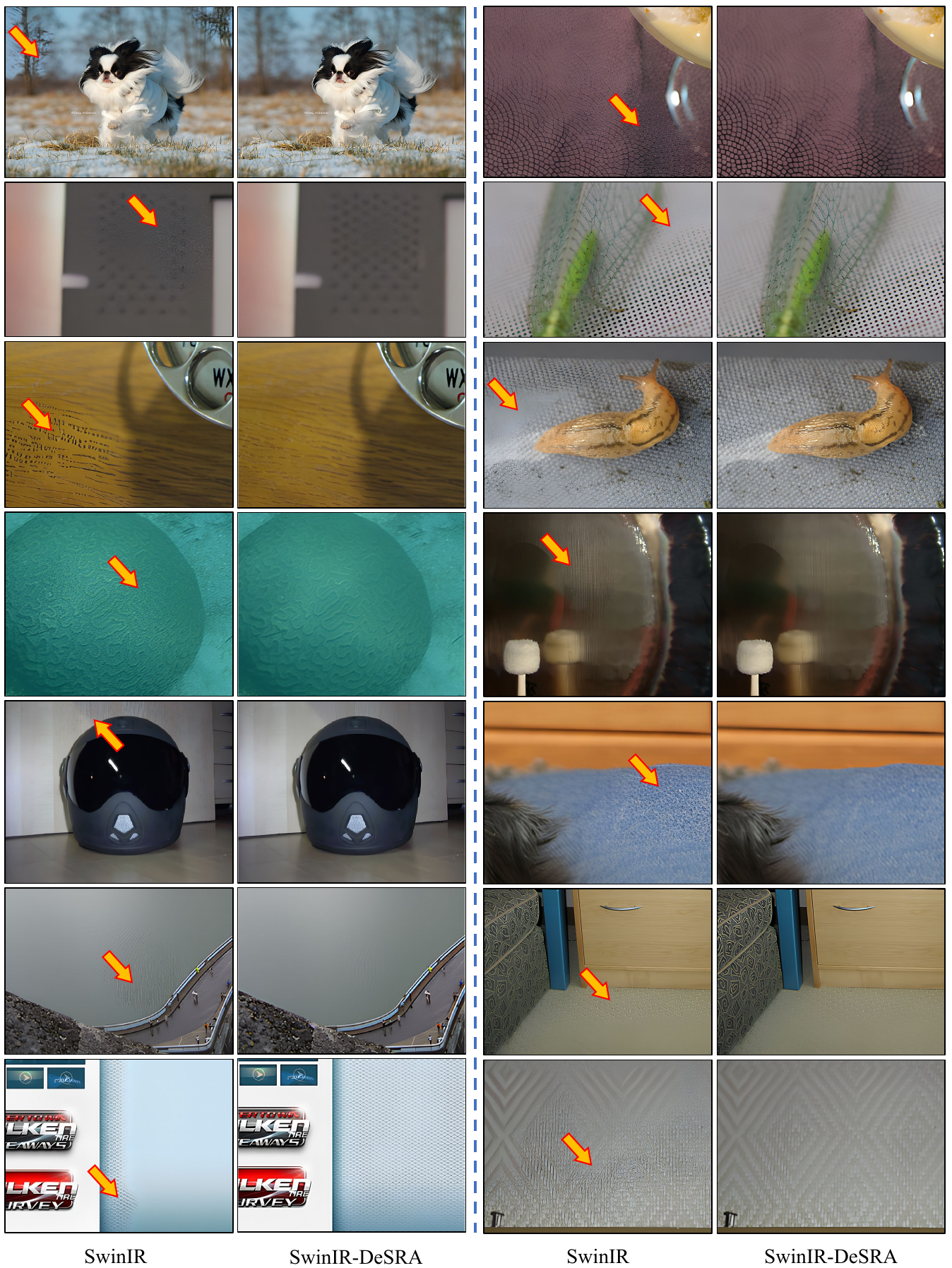}
		\caption{Visual comparison of results generated from SwinIR and SwinIR-DeSRA. Artifacts are obviously alleviated for results produced by SwinIR-DeSRA. \textbf{Zoom in for best view}}
		\label{fig:swinir_finetune}
	\end{center}
\end{figure*}

We provide the visual comparison between results with and without using our method to improve GAN-SR models, as shown in Fig.~\ref{fig:realesrgan_finetune}, Fig.~\ref{fig:ldl_finetune} and Fig.~\ref{fig:swinir_finetune}. We can observe that results generated by the improved GAN-SR models have greatly better visual quality
without obvious GAN-SR artifacts compared to the original inference results. All these experimental results demonstrate the effectiveness of our method for alleviating the artifacts and improving the GAN-SR model (Real-ESRGAN, LDL, and SwinIR).

\clearpage
\newpage
\subsection{Unrealiable of NIQE and MANIQA Metrics}
\label{no-reference-iqa}
\begin{figure*}[th]
	\begin{center}
		\includegraphics[width=0.8\linewidth]{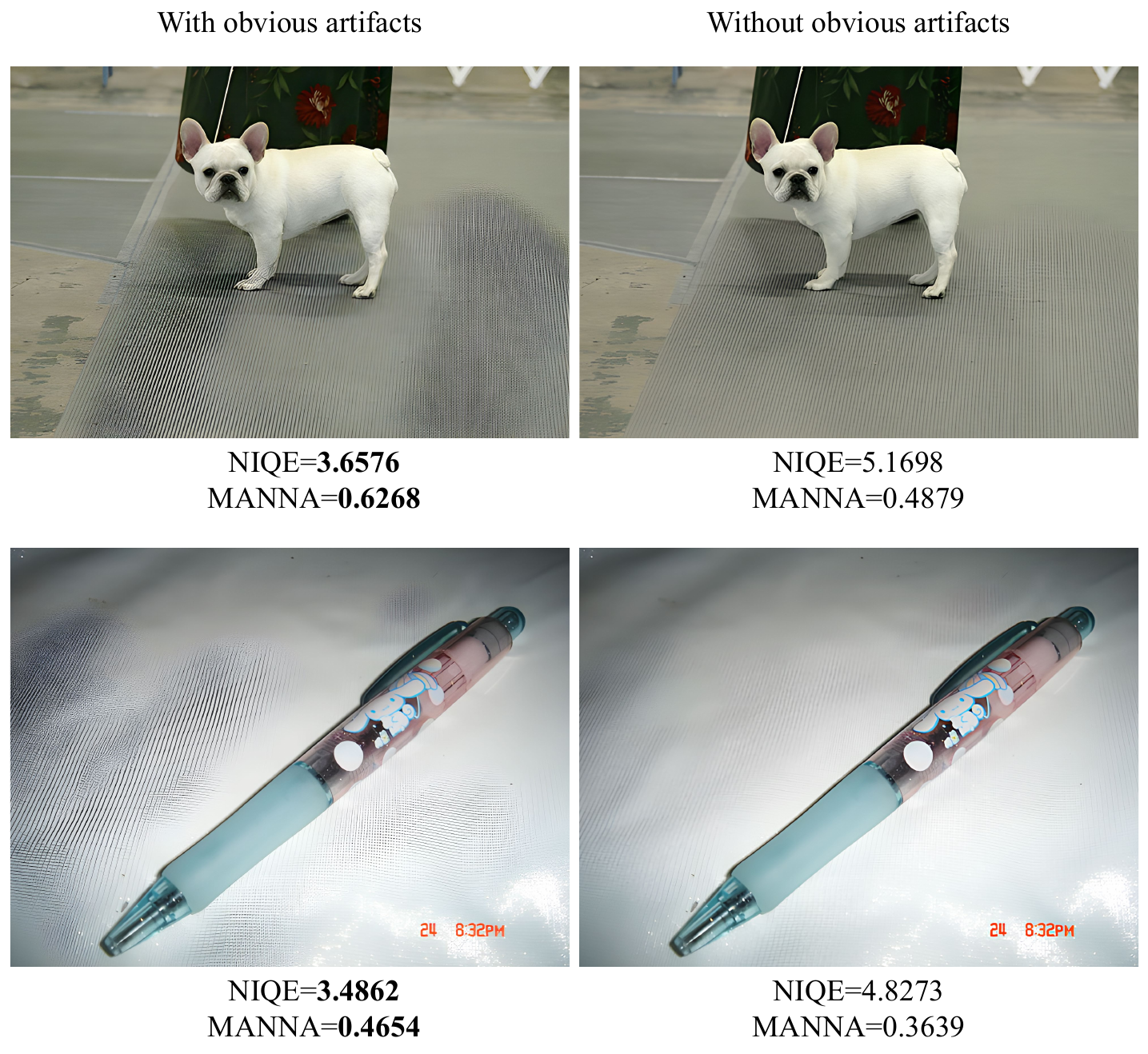}
		\caption{Evaluation of images with or without obvious artifacts on NIQE and MANNA metrics. Both of these two metrics cannot reflect the effects of artifact removal. \textbf{Zoom in for best view}}
		\label{fig:metrics}
	\end{center}
\end{figure*}

In Tab. 3 of the main paper and Tab.~\ref{tab:improved-swinir} in this supplementary material, we adopt IoU, Removal rate, and Addition rate metrics to evaluate the performance of improved GAN-SR models with DeSRA. Although NIQE~\cite{niqe} is the commonly-used metric in GAN-SR works, we observe that this metric cannot well reflect the performance of the improved GAN-SR models. As illustrated in Fig.~\ref{fig:metrics}, it can be obviously observed that the images in the second column have better visual results with fewer artifacts than the images in the first column. However, the values of NIQE (lower is better) and MANIQA~\cite{manna} (higher is better) show the opposite results.  MANIQA is the champion of the NTIRE 2022 Perceptual Image Quality Assessment Challenge. Therefore, we do not adopt these two metrics to evaluate the performance.

\end{document}